\documentclass{article} %
\usepackage{iclr2025_conference,times}

\usepackage{amsmath,amsfonts,bm}

\def\eqref#1{equation~\ref{#1}}

\def\1{\bm{1}}

\DeclareMathAlphabet{\mathsfit}{\encodingdefault}{\sfdefault}{m}{sl}
\SetMathAlphabet{\mathsfit}{bold}{\encodingdefault}{\sfdefault}{bx}{n}

\usepackage[utf8]{inputenc} %
\usepackage[T1]{fontenc}    %
\usepackage{hyperref}       %
\usepackage{url}            %
\usepackage{booktabs}       %
\usepackage{amsfonts}       %
\usepackage{nicefrac}       %

\usepackage{microtype}      %
\usepackage{xcolor}         %
\usepackage{hyperref}
\usepackage{url}
\usepackage{bm,mathtools}
\usepackage{multirow}
\usepackage{multicol}
\usepackage{floatrow}
\usepackage[capitalise]{cleveref}
\usepackage{float}
\floatstyle{plaintop}
\restylefloat{table}
\usepackage{caption}
\usepackage{enumitem}
\usepackage{booktabs}
\usepackage{graphicx}
\usepackage{natbib}
\usepackage{soul}

\title{SV4D: Dynamic 3D Content Generation with Multi-Frame and Multi-View Consistency}

\author{%
  Yiming Xie$^{1,2*}$
  \quad
  Chun-Han Yao$^{1*}$ \quad
  Vikram Voleti$^1$ \quad
  Huaizu Jiang$^{2\dagger}$ \quad
  Varun Jampani$^{1\dagger}$
  \\
  $^1$ Stability AI \quad
  $^2$ Northeastern University\\
  $^*$ Equal contribution \quad
  $^\dagger$ Equal advising
  \\
}

\newif\ifdrafting
\draftingtrue %
\ifdrafting
    \newcommand{\cy}[1]{\textcolor{blue}{CY: #1}}
    \newcommand{\yx}[1]{\textcolor{orange}{YX: #1}}
    \newcommand{\VJ}[1]{{\color{magenta}[VJ: #1]}}
    \newcommand{\todo}[1]{{\color{red}{#1}}}
    \newcommand{\newadd}[1]{\textcolor{black}{#1}}
\else
    \newcommand{\cy} [1] {}
    \newcommand{\yx} [1] {}
    \newcommand{\VJ} [1] {}
    \newcommand{\todo} [1] {}
    \newcommand{\newadd} [1] {}
\fi

\def\ie{\emph{i.e}}

\newcommand{\inlinesection}[1]{\vspace{1mm} \noindent {\bf #1}}

\newcommand{\urlNewWindow}[1]{\href[pdfnewwindow=true]{#1}{\nolinkurl{#1}}}

\newcommand{\Times}{{\mkern-2mu\times\mkern-2mu}}

\iclrfinalcopy %
\begin{document}

\maketitle

\begin{abstract}
We present Stable Video 4D (SV4D) — a latent video diffusion model for multi-frame and multi-view consistent dynamic 3D content generation.
Unlike previous methods that rely on separately trained generative models for video generation and novel view synthesis, we design a unified diffusion model to generate novel view videos of dynamic 3D objects. 
Specifically, given a monocular reference video, SV4D generates novel views for each video frame that are temporally consistent.
We then use the generated novel view videos to optimize an implicit 4D representation (dynamic NeRF) efficiently, without the need for cumbersome SDS-based optimization used in most prior works.
To train our unified novel view video generation model, we curate a dynamic 3D object dataset 
from the existing Objaverse dataset.
Extensive experimental results on multiple datasets and user studies demonstrate SV4D's state-of-the-art performance on novel-view video synthesis as well as 4D generation compared to prior works.
Project page: \urlNewWindow{https://sv4d.github.io}.
\end{abstract}
\vspace{-1em}

\section{Introduction}
The 3D world we live in is dynamic in nature with moving people, playing pets, bouncing balls, waving flags, etc.
Dynamic 3D object generation, also known as 4D generation, is the task of generating not just the 3D shape and appearance (texture) of a 3D object, but also its motion in 3D space. In this work, we tackle the problem of generating a 4D (dynamic 3D) object from a single monocular video of that object. 4D generation enables the effortless creation of realistic visual experiences, such as for video games, movies, AR/VR, etc.

4D generation from a single video is highly challenging as this involves simultaneously reasoning both the 
object appearance and motion at unseen camera views around the object.
This is also an ill-posed problem as a multitude of 4D results can plausibly explain a single given video.
There are two main technical challenges in training a robust 4D generative model that can generalize
to different object types and motions. 
First, there exists no large-scale dataset with 4D objects to train a robust generative model.
Second, the higher dimensional nature of the problem requires a large number of parameters to represent the 3D shape, appearance, and motion of an object.

\begin{figure}[t!]
    \centering
    \includegraphics[width=\linewidth]{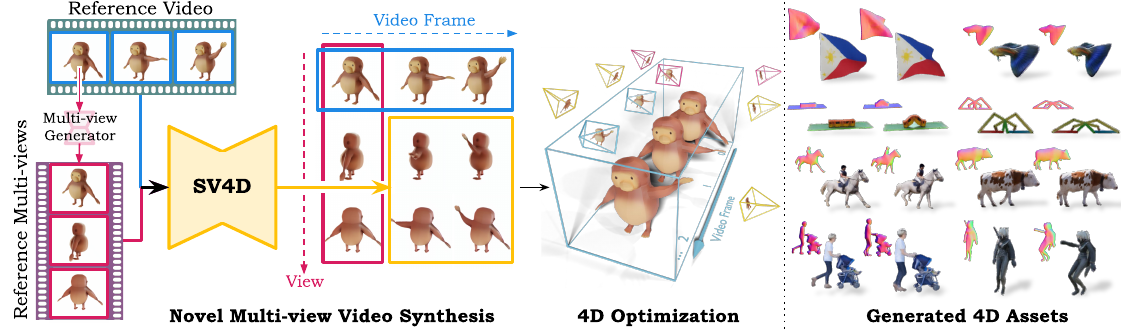}
    \caption{
        {\bf Stable Video 4D (SV4D)} framework overview and generated 4D assets. We adapt and train a video diffusion model to generate 
        novel view videos, conditioned on a single-view video and a multi-view orbital video of the first frame. 
        SV4D generated novel view videos are consistent across both view and motion axis, which we directly use to optimize dynamic 3D objects without the cumbersome SDS loss.
    }
\label{fig:pipeline}
\end{figure}

As a result, 
several recent techniques~\citep{singer2023text4d,bah20244dfy,ling2023alignyourgaussians,ren2023dreamgaussian4d,zhao2023animate124,jiang2023consistent4d,zeng2024stag4d,yin20234dgen} optimize 4D content by leveraging priors in pre-trained video and multi-view generative models
via score-distillation sampling (SDS) loss~\citep{poole2022dreamfusion} and its variants. %
However, they tend to produce unsatisfactory results due to independent modeling of object motion using video models, and novel view synthesis using multi-view generative models. In addition, they tend to take hours to generate a single 4D object due to time-consuming SDS-based optimization.
Concurrent works~\citep{yang2024diffusion,sun2024eg4d} try to partially address these issues by jointly sampling novel view videos (along both video frame and view axes) using both video
and multi-view
generative models; and then using the resulting novel view videos for 4D optimization. This results in a better novel view video synthesis with multi-view dynamic consistency, but several inconsistencies still remain due to the use of separate video and multi-view generative models.

In this work, we propose Stable Video 4D (SV4D) model that takes as input a single video, such as of a dynamic object, along with a user-specified camera trajectory around the object, and outputs videos of the object along each of the specified camera views. That is, given a video with $F$ number of frames and a camera trajectory with $V$ number of camera views, SV4D outputs a $V \Times F$ grid of images, as illustrated in~\cref{fig:pipeline}. Given a reference video, we first obtain the reference multi-views for the first frame using an off-the-shelf multi-view generator (SV3D~\citep{voleti2024sv3d}). SV4D then jointly outputs the remaining grid of images as highlighted by a yellow box in~\cref{fig:pipeline}.
In contrast to prior and concurrent works, SV4D jointly reasons along both view and motion axes, resulting in state-of-the-art multi-frame and multi-view consistency in the output novel view videos.

Specifically, we start with Stable Video Diffusion (SVD)~\citep{blattmann2023stable}, a state-of-the-art video generator, and equip it with two attention blocks: view attention and frame attention.
The view attention block aligns the multi-view images at each video frame, conditioning on the first view \ie, in the reference video.
Similarly, the frame attention block aligns the multi-frame images at each view, conditioning on the first frame at each view, \ie, reference multi-view.
This dual-attention design leads to significantly improved dynamic and multi-view consistency compared to prior art. 

A key challenge here is that SV4D needs to simultaneously generate the $V \Times F$ grid of images, which can quickly become large with long input videos; making it infeasible to fit into memory even on modern GPUs. As a remedy, we propose techniques to sequentially process an interleaved subset of input frames while also retaining consistency in the output image grid.
After generating the multiple novel view videos, we optimize a 4D representation of the dynamic 3D asset, as illustrated in~\cref{fig:pipeline}.

Given the lack of large-scale 4D datasets, we carefully initialized SV4D weights with those from SVD~\citep{blattmann2023stable} and SV3D~\citep{voleti2024sv3d} networks, thereby leveraging the priors learned in the existing video and multi-view diffusion models.
To further train SV4D, we carefully curated a subject of Objaverse~\citep{deitke2023objaverse,deitke2023objaversexl} dataset with dynamic 3D objects, resulting in the ObjaverseDy dataset.

We perform extensive comparisons of both novel view video synthesis and 4D generation results with respective state-of-the-art methods on datasets with synthetic (ObjaverseDy, Consistent4D~\citep{jiang2023consistent4d}), and real-world (DAVIS~\citep{Perazzi_CVPR_2016, Pont-Tuset_arXiv_2017,Caelles_arXiv_2019}) data.
We modify the FVD metric~\citep{unterthiner2018towards} 
to evaluate both video frame and view consistency,
validating the effectiveness and design choices of our approach. 
Fig.~\ref{fig:pipeline} shows some sample results of our approach. 

To summarize, our contributions include:
\begin{itemize}[leftmargin=*]
\vspace{-2mm}
    \item A novel SV4D network that can simultaneously reason across both frame and view axes. To our knowledge, this is the first work that trains a single novel view video synthesis network using 4D datasets, that can jointly perform novel view synthesis as well as video generation.
    \vspace{-1mm}
    \item A mixed sampling scheme that enables sequential processing of arbitrary long input videos while also retaining the multi-frame and multi-view consistency.
    \vspace{-1mm}
    \item State-of-the-art results on multiple benchmark datasets with both novel view video synthesis as well as 4D generation.
\end{itemize}

\vspace{-2mm}
\section{Related Work}
\vspace{-3mm}
\textbf{3D Generation}.
Here, we refer to the works that generate static 3D content as 3D generation.
DreamFusion~\citep{poole2022dreamfusion} first proposed to distill priors from the 2D diffusion model via SDS loss to optimize the 3D content from text or image.
Several subsequent works~\citep{yi2023gaussiandreamer,tang2023dreamgaussian,shi2023mvdream,wang2024prolificdreamer,li2023sweetdreamer,weng2023consistent123,pan2023enhancing,chen2023text,sun2023dreamcraft3d,sargent2023zeronvs,EnVision2023luciddreamer,zhou2023dreampropeller,guo2023stabledreamer} try to solve issues caused by the original SDS loss, such as multi-face Janus, slow generation speed, and over-saturated/smoothed generations.
Recent works~\citep{hong2023lrm,jiang2022LEAP,wang2023pf,zou2023triplane,wei2024meshlrm,tochilkin2024triposr} try to directly predict the 3D model of an object via a large reconstruction model.
Another approach~\citep{liu2023zero,liu2023syncdreamer,long2023wonder3d,voleti2024sv3d,ye2023consistent,karnewar2023holofusion,instant3d2023,shi2023toss,shi2023zero123++,wang2023imagedream,liu2024one,liu2023one2} to 3D generation is generating dense multi-view images with sufficient 3D consistency.
3D content is reconstructed based on the dense multi-view images.
We follow this strategy, but generate consistent multi-view videos (instead of images) and then reconstruct the 4D object.

\textbf{Video Generation}.
Recent video generation models~\citep{ho2022video,voleti2022mcvd,blattmann2023align,blattmann2023stable,he2022lvdm,singer2022make,guo2023animatediff} have shown very impressive performance with consistent geometry and realistic motions. 
Video generation models have good generalization capabilities, as they are trained on large-scale image and video data that are easier to collect than large-scale 3D or 4D data.
Hence, they are commonly used as foundation models for various generation tasks.
Well-trained video generative models 
have shown their potential to generate multi-view images as a 3D generator~\citep{voleti2024sv3d,chen2024v3d,han2024vfusion3d,kwak2024vivid,melas20243d}.
SV3D~\citep{voleti2024sv3d} adapts SVD~\citep{blattmann2023stable} to generate novel multiple views.
In this work, we leverage the pre-trained video generation model for 4D generation by adding an additional view attention layer to align the multiview images.

\textbf{4D Generation}.
On one hand, recent optimization-based methods~\citep{singer2023text4d,bah20244dfy,ling2023alignyourgaussians,ren2023dreamgaussian4d,zhao2023animate124,jiang2023consistent4d,zeng2024stag4d,yin20234dgen,chen2024ct4d} can generate 4D content by distilling pre-trained diffusion models in a 4D representation~\citep{cao2023hexplane,kerbl20233d,mildenhall2021nerf} via SDS loss~\cite{poole2022dreamfusion}.
However, they tend to take hours to generate 4D content due to time-consuming optimization.
On the other hand, photogrammetry-based methods~\citep{yang2024diffusion,pan2024fast,sun2024eg4d} mimic a 3D object capture pipeline by directly generating multi-frame multi-view images of a 4D content with dynamic and multi-view consistency and then directly reconstructing 4D representations with them. 
Although these methods have much faster speeds, inference-only pipelines are adopted due to the paucity of the 4D data, thus making the spatial-temporal consistency still unsatisfactory. 
\newadd{Moreover, some models like EG4D~\citep{sun2024eg4d} further rely on a third model (SDXL-Turbo~\citep{sauer2025adversarial}) to refine the 4D assets.}
In this work, we proposed a monolithic model to generate a more consistent image grid, and we used a carefully curated 4D dataset to train the model.
Recently, there has been a surge of interest in 4D generation, resulting in the development of
concurrent works~\citep{liang2024diffusion4d,zhang20244diffusion,li2024vividzoo,ren2024l4gm,wang2024vidu4d,jiang2024animate3d}.
Diffusion4D~\citep{liang2024diffusion4d} only generates diagonal images (space-time), by fine-tuning a 4D-aware video diffusion model with motion magnitude guidance.
Vivid-ZOO~\citep{li2024vividzoo} combines the 3D and video generative model and then finetunes alignment modules to mitigate the incompatibility between reused layers.
4Diffusion~\citep{zhang20244diffusion} inserts temporal layers to a 3D-aware diffusion model to generate novel view videos.
Both Vivid-ZOO and 4Diffusion freeze pre-trained weights, limiting their ability to fully capture spatial-temporal information. 
Additionally, they generate low-resolution images or a limited number of frames due to memory constraints. 
In contrast, SV4D finetunes all layers of our unified model, allowing for comprehensive spatial-temporal learning, and uses a mixed sampling scheme, enabling the generation of high-resolution and long novel views.

\vspace{-2mm}
\section{Method}
\vspace{-3mm}
Our main idea is to build multi-frame and multi-view consistency in a 4D object by surgically combining the frame-consistency in a video diffusion model, with the multi-view consistency in a multi-view diffusion model. In our case, we choose SVD~\citep{blattmann2023stable} and SV3D~\citep{voleti2024sv3d} as the video and multi-view diffusion models respectively, for the advantages reasoned below. However, it is to be noted that any choice of attention-based diffusion models should work.

\vspace{-2mm}
\subsection{Novel View Video Synthesis via SV4D}
\vspace{-2mm}

\inlinesection{Problem Setting.} 
Formally, we begin with a monocular input video ${\bf J} \in \mathbb{R}^{F \Times D}$ of a dynamic object of $F$ frames and $D \coloneq 3 \Times H \Times W$ dimensions, where $H$ and $W$ are the height and width of each frame. Our goal is to generate an image matrix ${\bf M} \in \mathbb{R}^{V \Times F \Times D}$ of the 4D object consisting of $V$ camera views and $F$ dynamic frames of each frame. Similar to SV3D~\citep{voleti2024sv3d}, the multi-view frames follow a camera pose trajectory $\bm{\pi} \in \mathbb{R}^{V \Times 2} = \{(e_v, a_v)\}_{v=1}^V$ as a sequence of $V$ tuples of elevation $e$ and azimuth angles $a$. 
We generate this image matrix by iteratively denoising samples from a learned conditional distribution $p({\bf M} | {\bf J}, \bm{\pi})$, parameterized by a 4D diffusion model. Due to memory limitation in generating $V \Times F$ images simultaneously, we break down the full generation process into multiple submatrix generation steps.

\begin{figure}[t!]
    \centering
    \includegraphics[width=\linewidth]{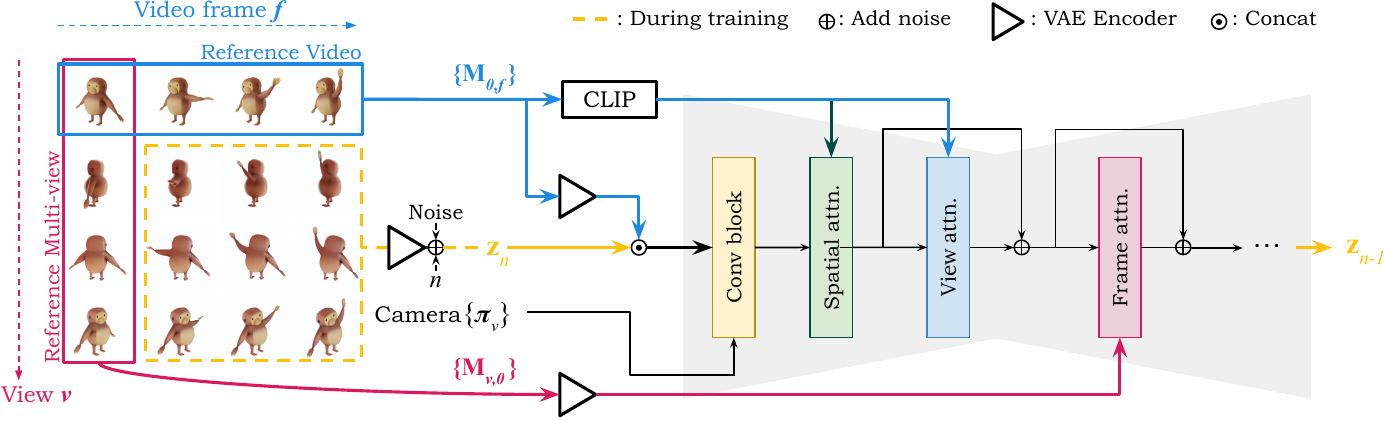}
    \caption{
        \textbf{\newadd{SV4D model architecture.}} For camera conditioning, we feed the sinusoidal embedding of camera viewpoints to the convolutional blocks in the UNet, and use the input video for cross-attention conditioning in the spatial and view attention blocks. To improve temporal consistency, we introduce an additional frame attention block, conditioned on the corresponding views of the first frame.
    }
\label{fig:architecture}
\end{figure}

\inlinesection{SV4D Network.}
Our goal is to make the generated image matrix $\bf M$ dynamically consistent in the ``\textbf{frame}'' axis, and multi-view consistent in the ``\textbf{view}'' axis (see \cref{fig:architecture} left).
To achieve this, we condition the image matrix generation on the corresponding \textbf{frames} of the monocular input video $\{{\bf M}_{0,f}\}={\bf J}$ as well as the \textbf{views} from reference multi-view images of the first video frame $\{{\bf M}_{v,0}\}$, which provide motion and multi-view information of the 4D object, respectively. 
Without loss of generality, we obtain the reference multi-view images of the first input frame by sampling from a pre-trained SV3D~\citep{voleti2024sv3d} model, which can be expressed as $p( \{{\bf M}_{v,0}\} | \{{\bf M}_{0,0}\} , \bm{\pi} )$. 
The overall SV4D novel view video synthesis can be rewritten as sampling from the distribution $p({\bf M} | \{{\bf M}_{0,f}\} , \{{\bf M}_{v,0}\} , \bm{\pi})$.

We build the SV4D network based on SVD~\citep{blattmann2023stable} and SV3D~\citep{voleti2024sv3d} models to combine the advantages of both video and multi-view diffusion models. 
As illustrated in~\cref{fig:architecture}, SV4D consists of a UNet with multiple layers, where each layer contains a sequence of one residual block with Conv3D layers and three transformer blocks with attention layers: spatial, view, and frame attention.
Similar to SV3D, the residual Conv block takes in the noisy latents of flattened image matrix as well as handles the incorporation of conditioning camera poses $\{\bm{\pi}_v\}$, and the spatial attention layer handles image-level details by performing attention across the image width and height axes.
To better capture motion cues in the input monocular video $\{{\bf M}_{0,f}\}$, we concatenate its VAE latents to the noisy latents ${\bf z}_n$ before feeding it to the UNet. 

For multi-view consistency, the view attention block transposes the features and performs attention in the multi-view axis. 
By using the CLIP embedding of corresponding input frames $\{{\bf M}_{0,f}\}$ as cross-attention conditioning, it allows the network to learn spatial consistency across novel views while maintaining the semantic context from the input video.

To further ensure dynamic consistency across video frames, we insert a frame attention layer in each UNet block, which applies the attention mechanism in the video frame dimension. 
The frame attention of each novel view video is conditioned on the corresponding reference view (of the first frame) $\{{\bf M}_{v,0}\}$ via cross-attention, allowing the network to preserve dynamic coherence starting from the first frame.
We initialize the weights of the frame attention layers from SVD and the rest of the network from SV3D$_p$, to leverage the generalizability as well as rich dynamic and multi-view priors learned from large-scale video and 3D datasets.
More network details are in Appendix~\ref{supp:network_details}.

\inlinesection{ObjaverseDy Dataset.}
Considering that there exists no large-scale training datasets with dynamic 3D objects, we curate a new 4D dataset from the existing Objeverse dataset~\citep{deitke2023objaverse,deitke2023objaversexl}, a massive dataset with annotated 3D objects.
Objaverse includes 
animated 3D objects, however, several of these animated 3D objects are not suitable for training due to having too few animated frames or insufficient motion.
In addition, in the rendering stage, the common rendering and sampling setting may cause some issues. 
For example, dynamic objects may be out of the image if the camera distance is fixed because they have global motion;
the motion of objects may be too fast or too slow if the temporal sampling step is fixed.

We follow several steps to curate and clean the 4D objects for our training purposes.
We first filter out the objects based on a review of licenses.
Then, we remove the objects whose animated frames are 
too few.
To further filter out objects with minimal motion, we subsample 
keyframes
from each video and apply simple thresholding on the maximum \textit{L1} distance between these frames as motion measurement. 
To render the training novel view videos, we flexibly choose the camera distance from the object. Starting from a base value, we increase the camera distance 
until the object fits within all frames of the rendered images. 
We also dynamically adjust the temporal sampling step.  
Starting from a base value, we increase the sampling step 
until the \textit{L1} distance between consecutive keyframes exceeds a certain threshold.
These steps ensure a high-quality collection of 4D objects, 
with rendered multi-view videos that form our ObjaverseDy dataset. 
More dataset details are in Appendix~\ref{supp:data_details}.

\inlinesection{Training Details.}
We train SV4D on our ObjaverseDy
dataset. 
We choose to finetune from the SV3D$_p$~\citep{voleti2024sv3d} model to output 40 frames ($F=5$ frames along each of the $V=8$ views) with the spatial resolution $576\times576$, where the parameters in the frame attention layers are initialized from SVD-xt~\citep{blattmann2023stable}.
Similar to SV3D, we train SV4D progressively by first training on the static camera orbits for 40K iterations, then fine-tuning it for 20K iterations on the dynamic orbits.
We use an effective batch size of 16 during training on 2 nodes of 8 80GB H100 GPUs. For more training details, please see Appendix~\ref{supp:training_details}.

\begin{figure}[t!]
    \centering
    \includegraphics[width=\linewidth]{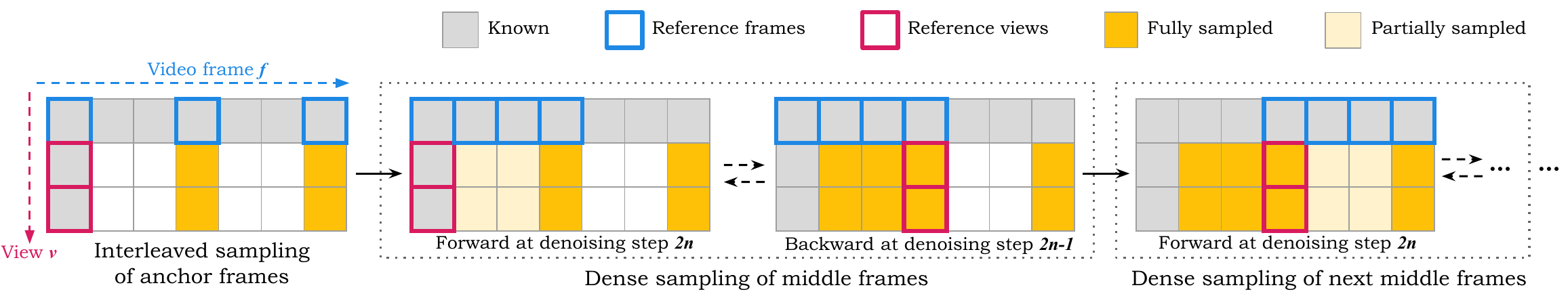}
    \caption{
        \textbf{SV4D model sampling} (7 frames and 3 views are for illustrative purposes only).
        To extend the generated multi-view videos while preserving temporal consistency, we propose a novel mixed-sampling strategy during inference. We first sample a sparse set of anchor frames, then use the anchor frames as new conditioning images to densely sample/interpolate the middle frames. To ensure a smooth transition between consecutive generations, we alternatively use the first (forward) or last (backward) frame within a time window for conditioning during dense sampling.
    }
\label{fig:sampling}
\end{figure}

\inlinesection{Inference Sampling Scheme.} 
Due to memory limitations, we cannot generate all the novel view frames at once.
To generate the full $V \Times F$ image matrix with arbitrary length videos,
one can naively run the submatrix generation independently.
However, we observe that it often leads to severe artifacts due to the inconsistencies between consecutive submatrices.
Hence, we design a novel sampling scheme to mitigate the issue. 
The proposed sampling scheme is illustrated in \cref{fig:sampling}.
Note that we only show motion frame extensions here for illustrative purposes.
We first generate a sparse set of anchor frames with SV4D (\textit{interleaved sampling}) as shown in \cref{fig:sampling} (left).
Then we use the anchor frames as new reference views to densely sample the remaining frames (\textit{dense sampling}).
To ensure a smooth transition between consecutive generations, we alternatively use the first (forward) or last (backward) anchor frame for conditioning at each diffusion step, as shown in \cref{fig:sampling} (middle).
In experiments (\cref{fig:sampling_compare}), we demonstrate that our sampling scheme can generate novel view videos that are more temporally consistent compared to using an off-the-shelf video interpolation model~\citep{licvpr23amt} to interpolate in between the SV4D-generated anchor frames.

\vspace{-2mm}
\subsection{4D Optimization from SV4D Generated Novel View Videos}
\vspace{-2mm}

\begin{figure}[t!]
    \centering
    \vspace{-2mm}
    \includegraphics[width=\linewidth]{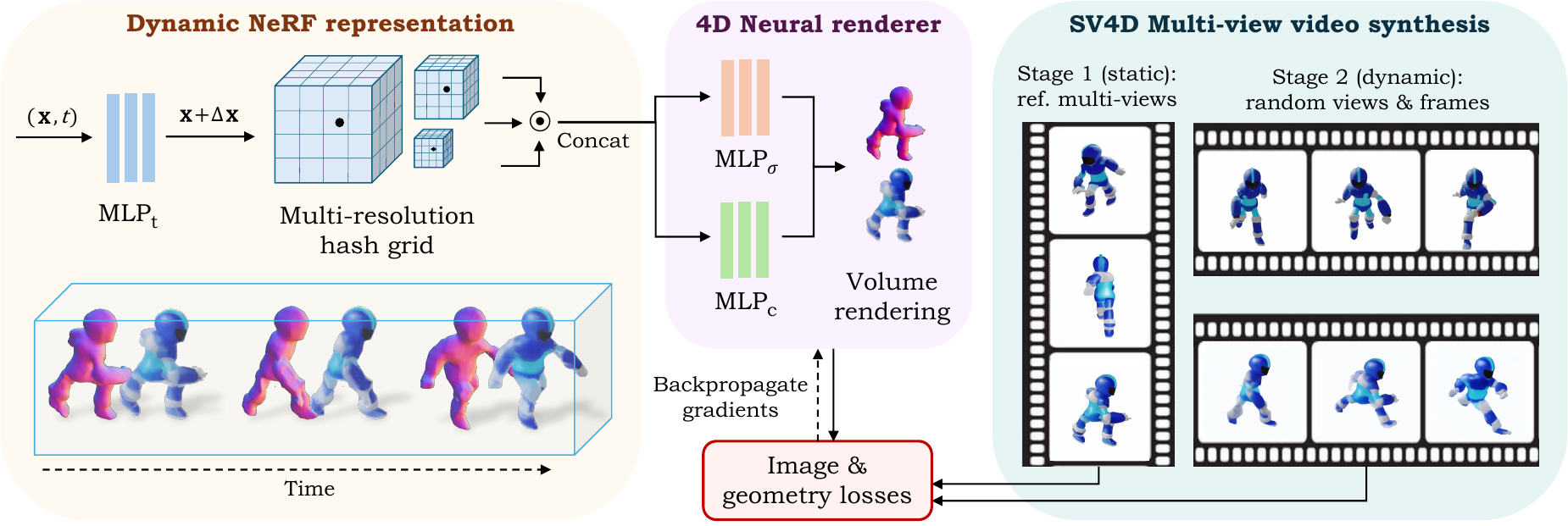}
    \caption{
        \textbf{Overview of optimization framework.}
        We first use the reference multi-view images (of the first frame) to optimize a static NeRF represented by a multi-resolution hash grid as well as density and color MLPs. Then, we unfreeze the temporal deformation MLP and optimize the dynamic NeRF with randomly sampled views and frames.
    }
\label{fig:4d_optimization}
\end{figure}

\inlinesection{4D Representation.} 
As shown in \cref{fig:4d_optimization}, given the novel view videos generated by SV4D, we optimize a 4D representation to reconstruct the dynamic 3D asset.
Formally, we learn a neural representation $\Psi_\theta: (\mathbf{x}, t) \mapsto (\sigma, c)$ that maps the sampled 3D points $\mathbf{x} = (x,y,z)$ along a camera ray, and the time embedding $t$ (i.e. continuous version of ``frame'' in the earlier discrete case) to its volumetric density $\sigma \in \mathbb{R}_+$ and color $c \in \mathbb{R}^3_+$.
Similar to D-NeRF~\citep{pumarola2021d}, we represent a 4D object by the composition of a canonical NeRF that captures the static 3D appearance and a deformation field which handles the object motion across time.
For each 3D point $\mathbf{x}$ in the canonical space, we trilinearly interpolate the multi-resolution hash grid features following Instant-NGP~\citep{mueller2022instant}, and decode them as density and color via MLP$_\mathrm{\sigma}$ and MLP$_\mathrm{c}$, respectively.
The deformation field is represented by an MLP network MLP$_\mathrm{t}$ conditioned on time embedding $t$, mapping temporal deformation of $\mathbf{x}$ to the common canonical space $\mathbf{x}+\Delta\mathbf{x}$.
Overall, the dynamic NeRF parameters $\theta$ include the canonical hash grid, MLP$_\mathrm{t}$, MLP$_\mathrm{\sigma}$, and MLP$_\mathrm{c}$. Since we do not exhaustively sample temporal timestamps or dense spatial views like in prior SDS-based approaches, we observe that this dynamic NeRF representation produces better 4D results compared to other representations such as 4D Gaussian Splatting~\citep{wu20244d}, which suffers from flickering artifacts and does not interpolate well across time or views.
\newadd{We show several examples to demonstrate this in the supplementary rebuttal video.}

\inlinesection{Optimization Details.} 
By leveraging the consistent image matrix generated by SV4D as pseudo ground-truths, we adopt a simple photometry-based optimization without the cumbersome SDS losses.
The reconstruction losses include a pixel-level MSE loss, mask loss, and a perceptual LPIPS~\citep{zhang2018unreasonable} loss.
Similar to SV3D~\citep{voleti2024sv3d}, we also use several geometric priors to regularize the output shapes, such as a mono normal
loss similar to MonoSDF~\citep{yu2022monosdf} as well as bilateral depth and normal smoothness losses~\citep{voleti2024sv3d} to encourage smooth 3D surfaces where the projected image gradients are low.

For training efficiency and stability, we follow a coarse-to-fine, static-to-dynamic strategy to optimize a 4D representation. That is, we first freeze the deformation field MLP$_\mathrm{t}$ and only optimize the canonical NeRF on the multi-view images of the first frame, while gradually increasing the rendering resolution from $128 \Times 128$ to $512 \Times 512$.
Then, we unfreeze MLP$_\mathrm{t}$ and randomly sample 4 frames $\Times$ 4 views for training.
Following the static-to-dynamic strategy, we also gradually optimize the time embedding $t$ from low to high temporal frequency.
In our experiments, we find that sampling more timestamps in one batch and progressive optimization techniques are crucial to 4D output quality.
We render the dynamic NeRF at $512\Times512$ resolution and use an Adam~\citep{kingma2014adam} optimizer to train all model parameters.
The overall optimization takes roughly 15-20 minutes per object. %
More implementation details can be found in Appendix~\ref{supp:opt_details}.

\vspace{-2mm}
\section{Experiments}
\vspace{-2mm}

\inlinesection{Datasets.}
We evaluate SV4D-synthesized novel view videos and 4D optimization results on the synthetic datasets ObjaverseDy and Consistent4D~\citep{deitke2023objaverse}.
Consistent4D dataset contains
dynamic objects collected from Objaverse~\citep{deitke2023objaverse,deitke2023objaversexl}. 
We excluded these 
objects from our training set to make a fair comparison.
We used the same input video and evaluated views as Consistent4D.
For the visual comparison, we also used single-view videos from the real-world videos dataset DAVIS~\citep{Perazzi_CVPR_2016, Pont-Tuset_arXiv_2017,Caelles_arXiv_2019}.

\inlinesection{Metrics.}
We use the SV4D model to generate multiple novel view videos corresponding to the trajectories of the ground truth camera in the evaluation datasets.
We compare each generated image with its corresponding ground-truth, in terms of Learned Perceptual Similarity (\textit{LPIPS}~\citep{zhang2018unreasonable}) and CLIP-score (\textit{CLIP-S}) to evaluate visual quality.
In addition, we evaluate the video coherence by reporting FVD~\citep{unterthiner2018towards}, a video-level metric commonly used in video generation tasks. 
We calculate FVD with different ways (see \cref{fig:metrics}):
\textit{FVD-F}: calculate FVD over frames at each view.
\textit{FVD-V}: calculate FVD over views at each frame.
\textit{FVD-Diag}: calculate FVD over the diagonal images of the image matrix.
\textit{FV4D}: calculate FVD over all images by scanning them in a bidirectional raster order.

\inlinesection{Baselines.}
For \textit{novel view video synthesis}, we compare SV4D with several recent methods capable of generating multiple novel view videos from a single-view video, including
SV3D~\citep{voleti2024sv3d}, Diffusion$^2$~\citep{yang2024diffusion}, STAG4D~\citep{zeng2024stag4d}. It is to be noted that all of these methods are inference-only techniques, and do not involve direct training in the 4D space like our method. 
We run SV3D to generate multi-view images for each video frame separately. 
STAG4D used Zero123++~\citep{shi2023zero123++} as the multi-view generator, which fixed the view angle, and hence the novel views generated from STAG4D cannot be changed to be consistent with the views evaluated.
We reproduced 
STAG4D with SV3D as the multi-view generator. SV3D has been shown to generate more consistent 3D results than Zero123++, so this serves as a stronger baseline.
For reference, we also compare SV4D with the concurrent work 4Diffusion~\citep{zhang20244diffusion}. 
We do not report the \textit{FVD-F} metric for 4Diffusion, as it only generates 8 frames, which is inconsistent with the number of frames used in our evaluation. 
Note that the \textit{FVD-F} metric is highly sensitive to the number of frames.
For \textit{4D generation}, we compare SV4D with other methods that can generate 4D representations, including Consistent4D~\citep{jiang2023consistent4d}, STAG4D~\citep{zeng2024stag4d}, 
4Diffusion~\citep{zhang20244diffusion},
DreamGaussian4D (DG4D)~\citep{ren2023dreamgaussian4d}, GaussianFlow~\citep{gao2024gaussianflow}, 4DGen~\citep{yin20234dgen}, Efficient4D~\citep{pan2024fast}.
More baseline details are in Appendix~\ref{supp:baseline_details}.
\begin{table}[t!]
\RawFloats
    \begin{minipage}[t]{0.49\linewidth}\centering
        \caption{
        \textbf{Evaluation of novel view video synthesis on the Consistent4D dataset.} SV4D can achieve better video frame consistency while maintaining comparable image quality. 
        $^\dagger$ Our reproduced version.}
        \label{tab:nvs-consistent4d}
        \resizebox{0.97\textwidth}{!}{
        \begin{tabular}{ l c c c c}
        \toprule 
         Model & LPIPS$\downarrow$ & CLIP-S$\uparrow$ & FVD-F$\downarrow$ \\
         \midrule
         SV3D~\citep{voleti2024sv3d} & \textbf{0.129} & 0.925 & 989.53\\
         4Diffusion~\citep{zhang20244diffusion} & 0.164 & 0.863 & -\\
         Diffusion$^2$~\citep{yang2024diffusion} & 0.189 & 0.907 & 1205.16 \\
         STAG4D$^\dagger$~\citep{zeng2024stag4d} & 0.131 & \textbf{0.929} & 861.88\\
         \midrule
         SV4D & \textbf{0.129} & \textbf{0.929} & \textbf{677.68}\\
        \bottomrule
        \end{tabular}
        }
    \end{minipage}\hfill
    \begin{minipage}[t]{0.49\linewidth}\centering
        \caption{
            \textbf{Evaluation of 4D outputs on the Consistent4D dataset.} 
            SV4D can achieve better visual quality and video frame smoothness. 
        }
        \label{tab:4d-consistent4d}
        \resizebox{1.0\textwidth}{!}{
        \begin{tabular}{ l c c c c }
        \toprule 
         Model & LPIPS$\downarrow$ & CLIP-S$\uparrow$ & FVD-F$\downarrow$ \\
         \midrule
         Consistent4D~\citep{jiang2023consistent4d} & 0.160 & 0.87 & 1133.93 \\
         STAG4D~\citep{zeng2024stag4d} & 0.126 & 0.91 & 992.21\\
         4Diffusion~\citep{zhang20244diffusion} & 0.165 & 0.88 & - \\
         Efficient4D~\citep{pan2024fast} & 0.130 & \textbf{0.92} & - \\
         4DGen~\citep{yin20234dgen} & 0.140 & 0.89 & -\\ 
         DG4D~\citep{ren2023dreamgaussian4d} & 0.160 & 0.87 & -\\
         GaussianFlow~\citep{gao2024gaussianflow} & 0.140 & 0.91 & - \\
         \midrule
         SV4D & \textbf{0.118} & \textbf{0.92} & \textbf{732.40}\\
        \bottomrule
        \end{tabular}
        }
\end{minipage}\hfill
\end{table}

\begin{table}[t!]
\caption{
\textbf{Evaluation of novel view video synthesis on the ObjaverseDy dataset.}
SV4D can achieve superior performance in both video frame and multi-view consistency.
$^\dagger$ Our reproduced version.}
\vspace{-.75em}
\label{tab:nvs-objaverse}
\centering
\resizebox{0.81\textwidth}{!}{
\begin{tabular}{ l c c c c c c c }
\toprule 
 Model & LPIPS$\downarrow$ & CLIP-S$\uparrow$ & FVD-F$\downarrow$ & FVD-V$\downarrow$ & FVD-Diag$\downarrow$ & FV4D$\downarrow$ \\
 \midrule
 SV3D~\citep{voleti2024sv3d} & \textbf{0.131} & \textbf{0.920} & 729.67 & 375.49 & 526.78 & 690.49\\
 Diffusion$^2$~\citep{yang2024diffusion} & 0.188 & 0.869 & 1048.47 & 564.80 & 938.98 & 1320.29\\
 STAG4D$^\dagger$~\citep{zeng2024stag4d} & 0.133 & 0.917 & 652.43 & 469.07 & 636.83 & 546.56\\
 \midrule
 SV4D & 0.136 & \textbf{0.920} & \textbf{585.09} & \textbf{331.94} & \textbf{503.02} & \textbf{470.46}\\
\bottomrule
\end{tabular}
}
\vspace{-1em}
\end{table}

\begin{table}[t!]
\caption{
\textbf{Evaluation of 4D outputs on the ObjaverseDy dataset.} 
SV4D consistently outperforms baselines in terms of all metrics, demonstrating superior performance in visual quality (\textit{LPIPS} and \textit{CLIP-S}), video frame consistency (\textit{FVD-F}), multi-view consistency (\textit{FVD-V}), and multi-view video consistency (\textit{FVD-Diag} and \textit{FV4D}).
}
\vspace{-.75em}
\label{tab:4d-objaverse}
\centering
\resizebox{0.92\textwidth}{!}{
\begin{tabular}{ l c c c c c c }
\toprule 
 Model & LPIPS$\downarrow$ & CLIP-S$\uparrow$ & FVD-F$\downarrow$ & FVD-V$\downarrow$ & FVD-Diag$\downarrow$ & FV4D$\downarrow$\\
 \midrule
 Consistent4D~\citep{jiang2023consistent4d} & 0.165 & 0.896 & 880.54 & 488.38 & 741.52 & 871.95\\
 STAG4D~\citep{zeng2024stag4d} & 0.158 & 0.860 & 929.10 & 453.62 & 663.50 & 1003.16 \\
 DreamGaussian4D~\citep{ren2023dreamgaussian4d} & 0.152 & 0.897 & 697.80 & 450.58 & 615.68 & 638.15 \\
 \midrule
 SV4D & \textbf{0.131} & \textbf{0.905} & \textbf{659.66} & \textbf{368.53} & \textbf{525.65} & \textbf{614.35} \\
\bottomrule
\end{tabular}
}
\end{table}

\vspace{-2mm}
\subsection{Quantitative Comparison}
\vspace{-2mm}
\inlinesection{Novel View Video Synthesis.}
We quantitatively compare our method with the baselines in terms of novel view video synthesis results.
\cref{tab:nvs-consistent4d} reports the comparisons in the Consistent4D dataset.
Due to the fact that the evaluated views are too sparse, we only report \textit{FVD-F} in the Consistent4D dataset.
Our method can achieve much better performance in terms of video frame consistency while maintaining comparable performance in terms of image quality.
In particular, our approach has a significant reduction of $31.5\%$ and $21.4\%$ in \textit{FVD-F} compared to SV3D and STAG4D, respectively.
\cref{tab:nvs-objaverse} reports the comparisons in the Objaverse dataset.
Our method can achieve much better video frame consistency as well as multi-view consistency while maintaining comparable performance in terms of image quality.
In particular, our approach has much lower \textit{FVD-F} compared to baseline methods, demonstrating our generated videos are much smoother. 
In addition, SV4D achieves better \textit{FVD-V} which shows better multi-view consistency.
Our method also surpasses the previous state-of-the-art methods in terms of \textit{FVD-Diag} and \textit{FV4D}, proving that the synthesized novel view videos have better video frame and multi-view consistency.

\inlinesection{4D Generation.}
We quantitatively compare our optimized 4D outputs with the baselines in Consistent4D and ObjaverseDy dataset, as shown in \cref{tab:4d-consistent4d,tab:4d-objaverse}, respectively.
Our method consistently outperforms baselines in terms of all metrics, demonstrating superior performance in visual quality (\textit{LPIPS} and \textit{CLIP-S}), motion smoothness (\textit{FVD-F}), multi-view smoothness (\textit{FVD-V}), and motion-multi-view joint smoothness (\textit{FVD-Diag} and \textit{FV4D}).

\begin{figure}[t!]
    \centering
    \includegraphics[width=\linewidth]{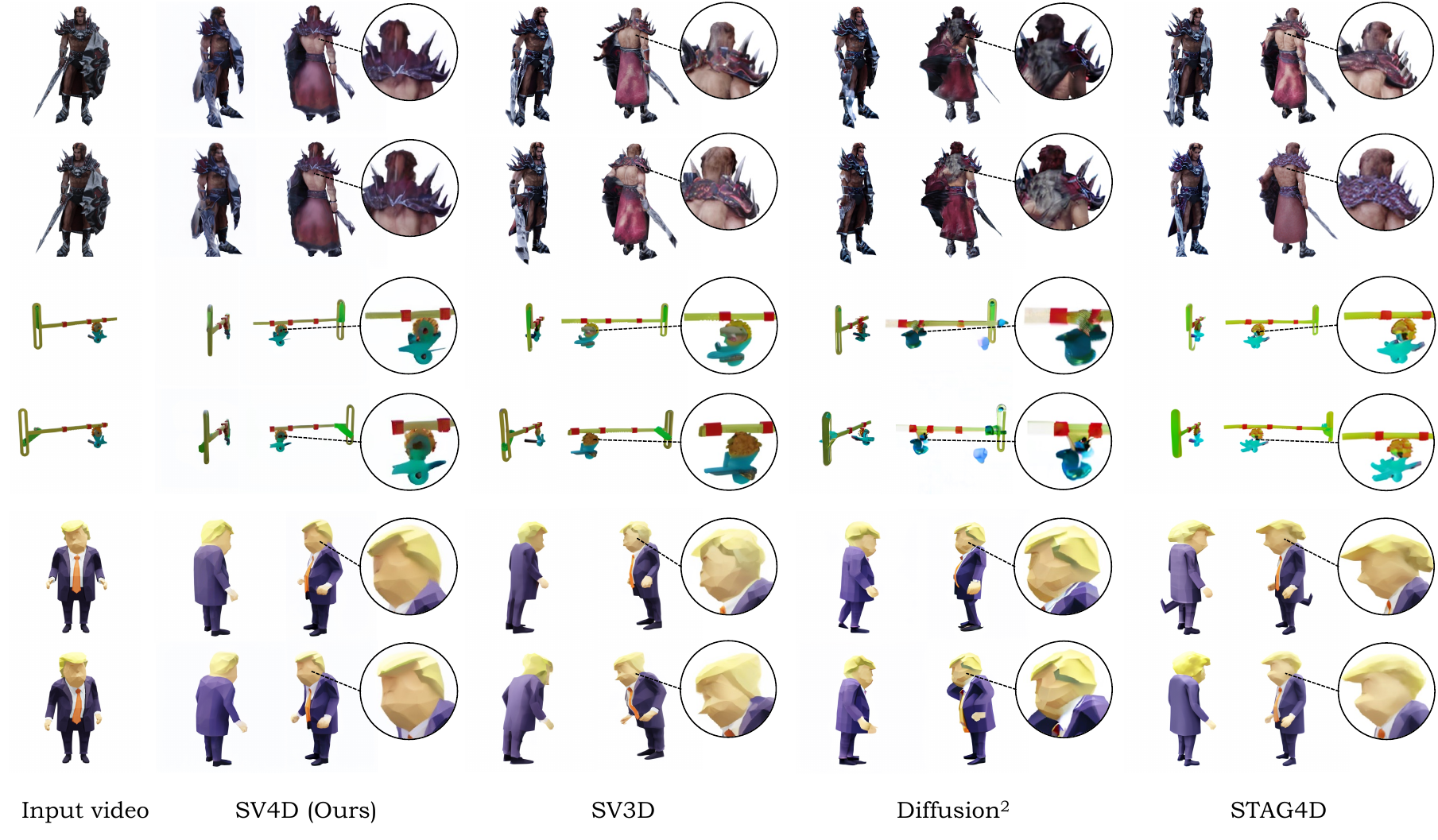}
    \vspace{-1.75em}
    \caption{
        \textbf{Visual comparison of novel view video synthesis results.}
        We show two frames in the input videos and two novel-view results of the corresponding frames. Compared to the baseline methods, SV4D outputs contain geometry and texture details that are more faithful to the input video and consistent across frames.
    }
\label{fig:nvs_results}
\end{figure}

\begin{figure}[t!]
    \centering
    \includegraphics[width=\linewidth]{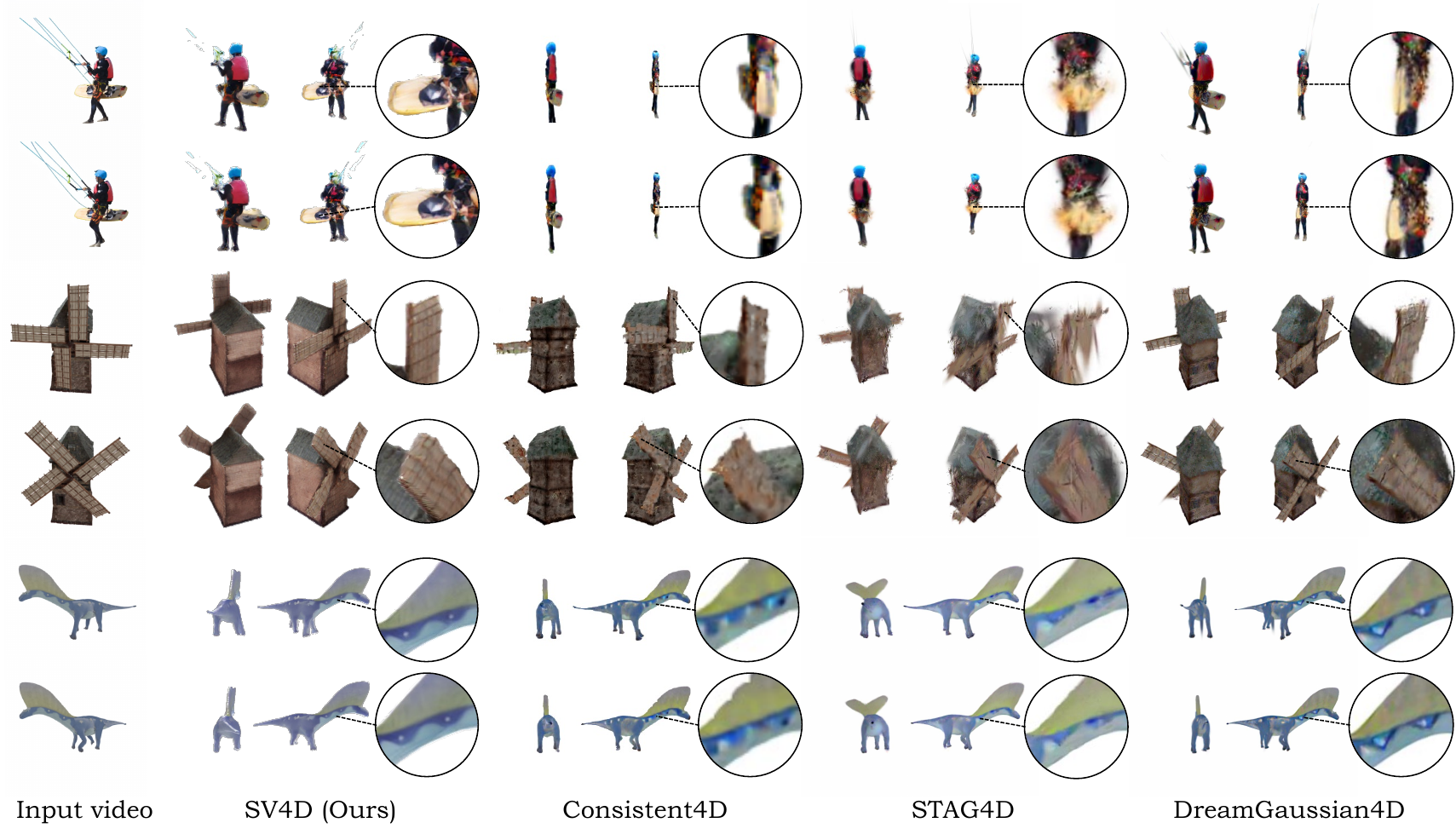}
    \vspace{-1.75em}
    \caption{
        \textbf{Visual comparison of generated 4D outputs.} 
        We show two frames in the input videos and render two novel views of the corresponding frames.
        By leveraging the consistent multi-view videos generated by SV4D, we can learn higher-quality 4D assets compared to the prior works with SDS-based losses. Our results are more detailed, consistent, and faithful to the input videos.
    }
\label{fig:4d_results}
\end{figure}

\vspace{-2mm}
\subsection{Visual Comparison}
\vspace{-2mm}
\inlinesection{Novel View Video Synthesis.}
In \cref{fig:nvs_results} we show the visual comparison of our multi-view video synthesis results against other methods. We observe that applying SV3D frame-by-frame leads to inconsistent geometry and texture in novel view videos. Diffusion$^2$ slightly improves the temporal coherence but tends to produce blurry or flickering artifacts. STAG4D can produce smoother videos but often fails at capturing large motion. Compared to these methods, SV4D can generate high-quality multi-view videos that are detailed, faithful to input videos, and temporally consistent.

\inlinesection{4D Generation.}
We compare our generated 4D results with the prior methods in \cref{fig:4d_results}. For each video, we render the 4D outputs at two different timestamps and two novel views. Consistent4D and STAG4D often produce blurry outputs with inconsistent geometry and texture. DreamGaussian4D can generate finer texture details but still suffers from flickering artifacts and sometimes creates inaccurate geometry. Moreover, all these methods rely on the computationally expensive SDS loss, which is prone to spatially incoherent results and over-saturated texture. On the contrary, SV4D optimizes 4D assets using purely photometric and geometric losses, resulting in smoother videos with realistic and faithful object appearance.
More visual comparison results can be found in Appendix~\ref{supp:additional_results}.

\vspace{-2mm}
\subsection{User Study}
\vspace{-2mm}
In addition to the quantitative evaluation, we also conduct two user studies on our multi-view video synthesis results and 4D outputs, respectively. Concretely, we randomly select 10 real-world videos from the DAVIS dataset and 10 synthetic videos from the Objaverse or Consistent4D datasets. For each video, we randomly choose a novel camera view and ask the user to compare the novel view videos generated by 4 different methods (SV4D and 3 baselines). The users are asked to select a video that ``looks more stable, realistic, and closely resembles the reference subject''. For multi-view video synthesis, SV4D results are preferred 73.3\% over per-frame SV3D (13.7\%), Diffusion$^2$ (5.0\%), and STAG4D (8.0\%) among multiple participants. For the optimized 4D outputs, SV4D achieves 60\% overall preference against Consistent4D (12.7\%), STAG4D (9.7\%), and DreamGaussian4D (17.6\%) among multiple users.

\begin{table}[t!]
\RawFloats
    \begin{minipage}[t]{0.49\linewidth}\centering
        \caption{
        \textbf{Evaluation of novel view video synthesis (anchor frames only)}.
        SV4D can effectively sample frames with faithful consistency and visual details.
        $\dagger$ Our reproduced version.
        }
        \label{tab:nvs_anchor}
        \resizebox{1.0\textwidth}{!}{
        \setlength{\tabcolsep}{1.8pt}
        \begin{tabular}{ l  c c  c c}
        \toprule 
         & \multicolumn{2}{c}{ObjaverseDy} & \multicolumn{2}{c}{Consistent4D}\\
         \cmidrule(l{2pt}r{2pt}){2-3}
         \cmidrule(l{2pt}r{2pt}){4-5}
         Model & FVD-F$\downarrow$ & FV4D$\downarrow$ & FVD-F$\downarrow$ & FV4D$\downarrow$\\
         \midrule
         SV3D~\citep{voleti2024sv3d} & 700.72 & 831.18 & 656.59 & 780.99\\
         Diffusion$^2$~\citep{yang2024diffusion} & 896.98 & 890.70 & 801.99 & 1093.24\\
         STAG4D$^\dagger$~\citep{zeng2024stag4d} & 765.79 & 669.28 & 601.23 & 659.79 \\
         \midrule
         SV4D & \textbf{629.22} & \textbf{569.08} & \textbf{469.73}& \textbf{621.00}\\
        \bottomrule
        \end{tabular}
        }
    \end{minipage}\hfill
    \begin{minipage}[t]{0.49\linewidth}\centering
        \caption{
        \textbf{Evaluation of different sampling strategies on ObjaverseDy dataset.}
        SV4D sampling can effectively generate full image matrixes with faithful consistency and visual details.
        }
        \label{tab:sampling}
        \resizebox{1.0\textwidth}{!}{
        \setlength{\tabcolsep}{0.2pt}
        \begin{tabular}{ l c c c c c c}
        \toprule 
         Sampling & LPIPS$\downarrow$ & FVD-F$\downarrow$ & FVD-Diag$\downarrow$ & FV4D$\downarrow$\\
         \midrule
         Interleaved & \textbf{0.136} & 731.28 & 567.15 & 717.21\\
         Independent & \textbf{0.136} & 663.41 & 512.14 & 488.31\\
         AMT~\citep{licvpr23amt} & 0.140 & 612.61 & 505.36 & 472.87\\
         \midrule
         SV4D & \textbf{0.136} & \textbf{585.09} & \textbf{503.02} & \textbf{470.46}\\
        \bottomrule
        \end{tabular}
        }
\end{minipage}\hfill
\end{table}

\begin{figure}[t]
    \begin{floatrow}
    \ffigbox{
        \includegraphics[width=0.7\linewidth]{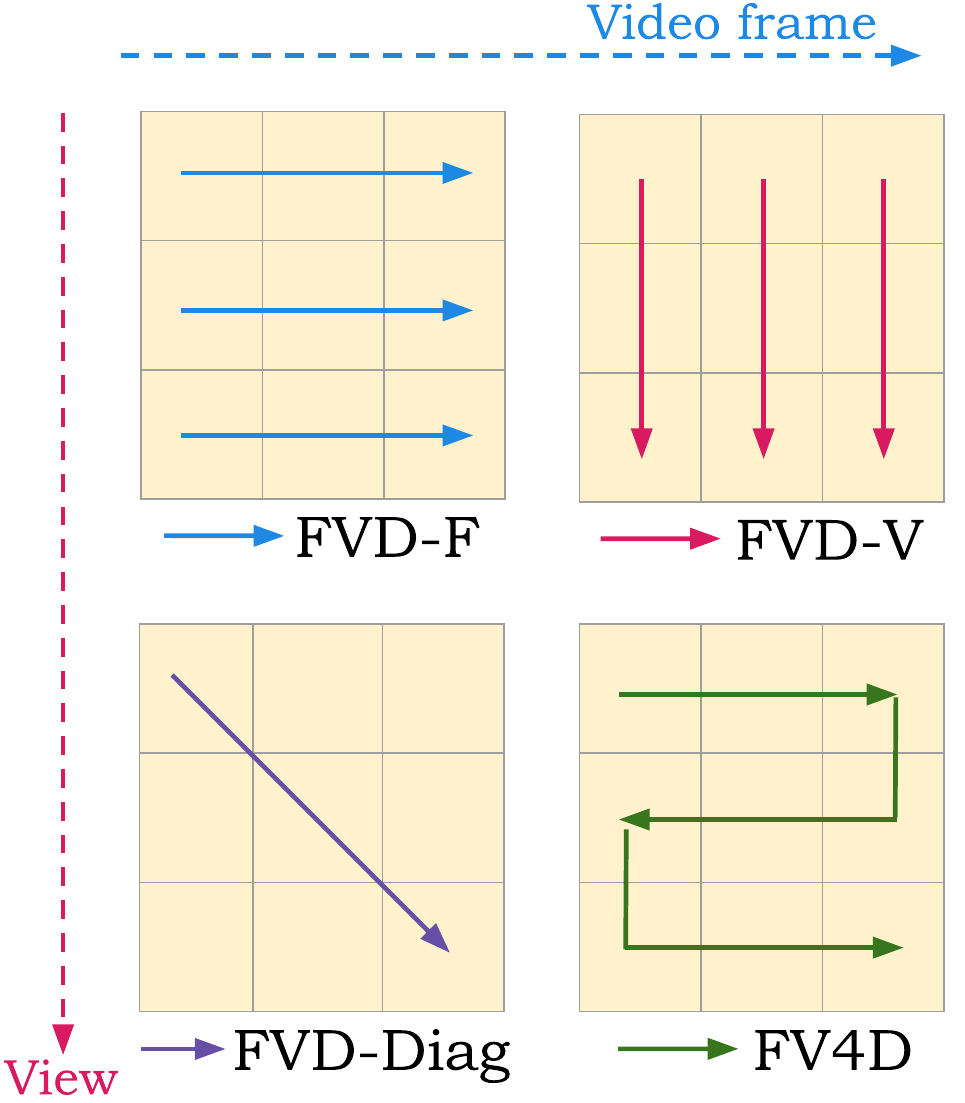}
    }{
      \caption{
        \textbf{Illustrations of video and 4D metrics.} 
        FVD-F evaluates coherence between video frames from a fixed view. FVD-V captures multi-view consistency. We also design FVD-Diag and FV4D to evaluate 4D consistency by traversing the image matrix through different paths.
        }
        \label{fig:metrics}
    }
    \ffigbox{
      \includegraphics[width=1.0\linewidth]{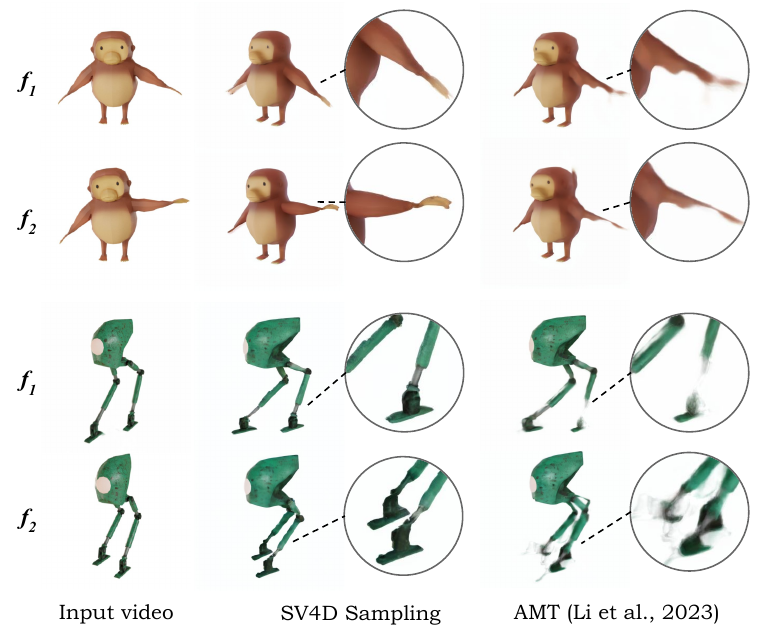}
      \vspace{-1.5em}
    }{
      \caption{
      \textbf{Visual comparison of SV4D sampling and video interpolation.} Given the same sparse anchor frames, SV4D can effectively sample the middle frames with faithful motion and details, whereas an off-the-shelf video interpolation model creates blurry results or missing parts.
      }
      \label{fig:sampling_compare}
    }
    \end{floatrow}
\end{figure}

\vspace{-2mm}
\subsection{Ablative Analyses}
\vspace{-2mm}
We conduct several ablation experiments to validate the effectiveness of our model's design choices. 
We summarize the key findings below.

\inlinesection{SV4D can generate anchor frames with better consistency.}
To validate the quality of anchor frames, we compare the anchor frames (8 views $\Times$ 5 video frames) generated from SV4D and the baselines on Consistent4D and ObjaverseDy dataset, as shown in \cref{tab:nvs_anchor}.
SV4D results have much lower \textit{FVD-F} and \textit{FV4D}, showing much better video frame and multi-view consistency.

\inlinesection{SV4D Sampling is better than off-the-shelf interpolation method.}
Extending anchor frames to the full $V \Times F$ image matrix is not trivial.
In \cref{tab:sampling}, we compare our SV4D sampling (\textit{ours}) and the off-the-shelf interpolation method AMT~\citep{licvpr23amt}.
The model with SV4D sampling outperforms AMT across all metrics. 
\cref{fig:sampling_compare} validates this observation, where we can see that the SV4D sampling can synthesize distinct images while the results of the off-the-shelf interpolation method have a lot of blur results or missing parts.
Table~\ref{tab:sampling} also shows the results of two intuitive sampling strategies, independent and interleaved sampling, both of which produce videos with lower consistency.
Appendix~\ref{supp:sampling_details} explains the details of independent and interleaved sampling.
These results validate that the SV4D sampling is effective.

\vspace{-2mm}
\section{Conclusion}
\vspace{-2mm}

We present SV4D, a latent video diffusion model for novel view video synthesis and 4D generation. Given an input video, SV4D can generate multiple novel view videos that are dynamically and spatially consistent, by leveraging the dynamic prior in SVD and multi-view prior in SV3D within a unified architecture. The generated novel view videos can then be used to efficiently optimize a 4D asset without SDS losses from one or multiple diffusion models. Our extensive experiments show that SV4D outputs are more multi-frame and multi-view consistent than existing methods and generalizable to real-world videos, achieving the state-of-the-art performance on novel view video synthesis and 4D generation. We believe that SV4D provides a solid foundation model for further research on dynamic 3D object generation.

\bibliography{iclr2025_conference}

\begin{thebibliography}{88}
\providecommand{\natexlab}[1]{#1}
\providecommand{\url}[1]{\texttt{#1}}
\expandafter\ifx\csname urlstyle\endcsname\relax
  \providecommand{\doi}[1]{doi: #1}\else
  \providecommand{\doi}{doi: \begingroup \urlstyle{rm}\Url}\fi

\bibitem[Bahmani et~al.(2024)Bahmani, Skorokhodov, Rong, Wetzstein, Guibas, Wonka, Tulyakov, Park, Tagliasacchi, and Lindell]{bah20244dfy}
Sherwin Bahmani, Ivan Skorokhodov, Victor Rong, Gordon Wetzstein, Leonidas Guibas, Peter Wonka, Sergey Tulyakov, Jeong~Joon Park, Andrea Tagliasacchi, and David~B. Lindell.
\newblock {4D-fy}: Text-to-4{D} generation using hybrid score distillation sampling.
\newblock In \emph{Proceedings of the IEEE/CVF Conference on Computer Vision and Pattern Recognition}, 2024.

\bibitem[Blattmann et~al.(2023{\natexlab{a}})Blattmann, Dockhorn, Kulal, Mendelevitch, Kilian, Lorenz, Levi, English, Voleti, Letts, et~al.]{blattmann2023stable}
Andreas Blattmann, Tim Dockhorn, Sumith Kulal, Daniel Mendelevitch, Maciej Kilian, Dominik Lorenz, Yam Levi, Zion English, Vikram Voleti, Adam Letts, et~al.
\newblock Stable video diffusion: Scaling latent video diffusion models to large datasets.
\newblock \emph{arXiv preprint arXiv:2311.15127}, 2023{\natexlab{a}}.

\bibitem[Blattmann et~al.(2023{\natexlab{b}})Blattmann, Rombach, Ling, Dockhorn, Kim, Fidler, and Kreis]{blattmann2023align}
Andreas Blattmann, Robin Rombach, Huan Ling, Tim Dockhorn, Seung~Wook Kim, Sanja Fidler, and Karsten Kreis.
\newblock Align your latents: High-resolution video synthesis with latent diffusion models.
\newblock In \emph{Proceedings of the IEEE/CVF Conference on Computer Vision and Pattern Recognition}, 2023{\natexlab{b}}.

\bibitem[Boss et~al.(2022)Boss, Engelhardt, Kar, Li, Sun, Barron, Lensch, and Jampani]{boss2022-samurai}
Mark Boss, Andreas Engelhardt, Abhishek Kar, Yuanzhen Li, Deqing Sun, Jonathan~T. Barron, Hendrik~P.A. Lensch, and Varun Jampani.
\newblock {SAMURAI}: {S}hape {A}nd {M}aterial from {U}nconstrained {R}eal-world {A}rbitrary {I}mage collections.
\newblock In \emph{Advances in Neural Information Processing Systems (NeurIPS)}, 2022.

\bibitem[Caelles et~al.(2019)Caelles, Pont-Tuset, Perazzi, Montes, Maninis, and Van~Gool]{Caelles_arXiv_2019}
Sergi Caelles, Jordi Pont-Tuset, Federico Perazzi, Alberto Montes, Kevis-Kokitsi Maninis, and Luc Van~Gool.
\newblock The 2019 davis challenge on vos: Unsupervised multi-object segmentation.
\newblock \emph{arXiv preprint arXiv:1905.00737}, 2019.

\bibitem[Cao \& Johnson(2023)Cao and Johnson]{cao2023hexplane}
Ang Cao and Justin Johnson.
\newblock Hexplane: A fast representation for dynamic scenes.
\newblock In \emph{Proceedings of the IEEE/CVF Conference on Computer Vision and Pattern Recognition}, 2023.

\bibitem[Chen et~al.(2024{\natexlab{a}})Chen, Huang, Chen, Chen, Han, Zhang, and Gong]{chen2024ct4d}
Ce~Chen, Shaoli Huang, Xuelin Chen, Guangyi Chen, Xiaoguang Han, Kun Zhang, and Mingming Gong.
\newblock Ct4d: Consistent text-to-4d generation with animatable meshes.
\newblock \emph{arXiv preprint arXiv:2408.08342}, 2024{\natexlab{a}}.

\bibitem[Chen et~al.(2024{\natexlab{b}})Chen, Wang, Wang, and Liu]{chen2023text}
Zilong Chen, Feng Wang, Yikai Wang, and Huaping Liu.
\newblock Text-to-3d using gaussian splatting.
\newblock In \emph{Proceedings of the IEEE/CVF conference on computer vision and pattern recognition}, pp.\  21401--21412, 2024{\natexlab{b}}.

\bibitem[Chen et~al.(2024{\natexlab{c}})Chen, Wang, Wang, Wang, and Liu]{chen2024v3d}
Zilong Chen, Yikai Wang, Feng Wang, Zhengyi Wang, and Huaping Liu.
\newblock {V3D}: Video diffusion models are effective 3{D} generators.
\newblock \emph{arXiv preprint arXiv:2403.06738}, 2024{\natexlab{c}}.

\bibitem[Deitke et~al.(2023{\natexlab{a}})Deitke, Liu, Wallingford, Ngo, Michel, Kusupati, Fan, Laforte, Voleti, Gadre, et~al.]{deitke2023objaversexl}
Matt Deitke, Ruoshi Liu, Matthew Wallingford, Huong Ngo, Oscar Michel, Aditya Kusupati, Alan Fan, Christian Laforte, Vikram Voleti, Samir~Yitzhak Gadre, et~al.
\newblock Objaverse-xl: A universe of 10m+ 3d objects.
\newblock \emph{Advances in Neural Information Processing Systems}, 36:\penalty0 35799--35813, 2023{\natexlab{a}}.

\bibitem[Deitke et~al.(2023{\natexlab{b}})Deitke, Schwenk, Salvador, Weihs, Michel, VanderBilt, Schmidt, Ehsani, Kembhavi, and Farhadi]{deitke2023objaverse}
Matt Deitke, Dustin Schwenk, Jordi Salvador, Luca Weihs, Oscar Michel, Eli VanderBilt, Ludwig Schmidt, Kiana Ehsani, Aniruddha Kembhavi, and Ali Farhadi.
\newblock Objaverse: A universe of annotated 3{D} objects.
\newblock In \emph{Proceedings of the IEEE/CVF Conference on Computer Vision and Pattern Recognition}, 2023{\natexlab{b}}.

\bibitem[Eftekhar et~al.(2021)Eftekhar, Sax, Bachmann, Malik, and Zamir]{Eftekhar2021omnidata}
Ainaz Eftekhar, Alexander Sax, Roman Bachmann, Jitendra Malik, and Amir Zamir.
\newblock Omnidata: A scalable pipeline for making multi-task mid-level vision datasets from 3{D} scans.
\newblock In \emph{IEEE International Conference on Computer Vision (ICCV)}, 2021.

\bibitem[Gao et~al.(2024)Gao, Xu, Cao, Mildenhall, Ma, Chen, Tang, and Neumann]{gao2024gaussianflow}
Quankai Gao, Qiangeng Xu, Zhe Cao, Ben Mildenhall, Wenchao Ma, Le~Chen, Danhang Tang, and Ulrich Neumann.
\newblock {GaussianFlow}: Splatting gaussian dynamics for 4{D} content creation.
\newblock \emph{arXiv preprint arXiv:2403.12365}, 2024.

\bibitem[Guo et~al.(2023)Guo, Hao, Caccavale, Ren, Zhang, Shan, Sankar, Schwing, Colburn, and Ma]{guo2023stabledreamer}
Pengsheng Guo, Hans Hao, Adam Caccavale, Zhongzheng Ren, Edward Zhang, Qi~Shan, Aditya Sankar, Alexander~G Schwing, Alex Colburn, and Fangchang Ma.
\newblock {StableDreamer}: Taming noisy score distillation sampling for text-to-3d.
\newblock \emph{arXiv preprint arXiv:2312.02189}, 2023.

\bibitem[Guo et~al.(2024)Guo, Yang, Rao, Liang, Wang, Qiao, Agrawala, Lin, and Dai]{guo2023animatediff}
Yuwei Guo, Ceyuan Yang, Anyi Rao, Zhengyang Liang, Yaohui Wang, Yu~Qiao, Maneesh Agrawala, Dahua Lin, and Bo~Dai.
\newblock Animatediff: Animate your personalized text-to-image diffusion models without specific tuning.
\newblock In \emph{The Twelfth International Conference on Learning Representations}, 2024.

\bibitem[Han et~al.(2024)Han, Kokkinos, and Torr]{han2024vfusion3d}
Junlin Han, Filippos Kokkinos, and Philip Torr.
\newblock Vfusion3d: Learning scalable 3d generative models from video diffusion models.
\newblock In \emph{European Conference on Computer Vision}, pp.\  333--350. Springer, 2024.

\bibitem[He et~al.(2022)He, Yang, Zhang, Shan, and Chen]{he2022lvdm}
Yingqing He, Tianyu Yang, Yong Zhang, Ying Shan, and Qifeng Chen.
\newblock Latent video diffusion models for high-fidelity long video generation.
\newblock \emph{arXiv}, 2022.

\bibitem[Ho et~al.(2022)Ho, Salimans, Gritsenko, Chan, Norouzi, and Fleet]{ho2022video}
Jonathan Ho, Tim Salimans, Alexey Gritsenko, William Chan, Mohammad Norouzi, and David~J Fleet.
\newblock Video diffusion models.
\newblock \emph{Advances in Neural Information Processing Systems}, 2022.

\bibitem[Hong et~al.(2024)Hong, Zhang, Gu, Bi, Zhou, Liu, Liu, Sunkavalli, Bui, and Tan]{hong2023lrm}
Yicong Hong, Kai Zhang, Jiuxiang Gu, Sai Bi, Yang Zhou, Difan Liu, Feng Liu, Kalyan Sunkavalli, Trung Bui, and Hao Tan.
\newblock Lrm: Large reconstruction model for single image to 3d.
\newblock In \emph{The Twelfth International Conference on Learning Representations}, 2024.

\bibitem[Jiang et~al.(2024{\natexlab{a}})Jiang, Jiang, Zhao, and Huang]{jiang2022LEAP}
Hanwen Jiang, Zhenyu Jiang, Yue Zhao, and Qixing Huang.
\newblock Leap: Liberate sparse-view 3d modeling from camera poses.
\newblock In \emph{The Twelfth International Conference on Learning Representations}, 2024{\natexlab{a}}.

\bibitem[Jiang et~al.(2024{\natexlab{b}})Jiang, Yu, Cao, Wang, Hu, and Gao]{jiang2024animate3d}
Yanqin Jiang, Chaohui Yu, Chenjie Cao, Fan Wang, Weiming Hu, and Jin Gao.
\newblock Animate3d: Animating any 3d model with multi-view video diffusion.
\newblock \emph{arXiv}, 2024{\natexlab{b}}.

\bibitem[Jiang et~al.(2024{\natexlab{c}})Jiang, Zhang, Gao, Hu, and Yao]{jiang2023consistent4d}
Yanqin Jiang, Li~Zhang, Jin Gao, Weiming Hu, and Yao Yao.
\newblock Consistent4d: Consistent 360° dynamic object generation from monocular video.
\newblock In \emph{The Twelfth International Conference on Learning Representations}, 2024{\natexlab{c}}.

\bibitem[Kanopoulos et~al.(1988)Kanopoulos, Vasanthavada, and Baker]{kanopoulos1988sobel}
Nick Kanopoulos, Nagesh Vasanthavada, and Robert~L Baker.
\newblock Design of an image edge detection filter using the sobel operator.
\newblock \emph{IEEE Journal of solid-state circuits}, 23\penalty0 (2):\penalty0 358--367, 1988.

\bibitem[Karnewar et~al.(2023)Karnewar, Mitra, Vedaldi, and Novotny]{karnewar2023holofusion}
Animesh Karnewar, Niloy~J Mitra, Andrea Vedaldi, and David Novotny.
\newblock {HoloFusion}: Towards photo-realistic 3{D} generative modeling.
\newblock In \emph{Proceedings of the IEEE/CVF International Conference on Computer Vision}, 2023.

\bibitem[Karras et~al.(2022)Karras, Aittala, Aila, and Laine]{karras2022elucidating}
Tero Karras, Miika Aittala, Timo Aila, and Samuli Laine.
\newblock Elucidating the design space of diffusion-based generative models.
\newblock \emph{Advances in neural information processing systems}, 2022.

\bibitem[Kerbl et~al.(2023)Kerbl, Kopanas, Leimk{\"u}hler, and Drettakis]{kerbl20233d}
Bernhard Kerbl, Georgios Kopanas, Thomas Leimk{\"u}hler, and George Drettakis.
\newblock 3{D} gaussian splatting for real-time radiance field rendering.
\newblock \emph{ACM Transactions on Graphics}, 2023.

\bibitem[Kingma \& Ba(2014)Kingma and Ba]{kingma2014adam}
Diederik~P Kingma and Jimmy Ba.
\newblock Adam: A method for stochastic optimization.
\newblock \emph{arXiv preprint arXiv:1412.6980}, 2014.

\bibitem[Kwak et~al.(2024)Kwak, Dong, Jin, Ko, Mahajan, and Yi]{kwak2024vivid}
Jeong-gi Kwak, Erqun Dong, Yuhe Jin, Hanseok Ko, Shweta Mahajan, and Kwang~Moo Yi.
\newblock Vivid-1-to-3: Novel view synthesis with video diffusion models.
\newblock In \emph{Proceedings of the IEEE/CVF Conference on Computer Vision and Pattern Recognition}, 2024.

\bibitem[Li et~al.(2025)Li, Zheng, Zhu, Mai, Zhang, Wonka, and Ghanem]{li2024vividzoo}
Bing Li, Cheng Zheng, Wenxuan Zhu, Jinjie Mai, Biao Zhang, Peter Wonka, and Bernard Ghanem.
\newblock Vivid-zoo: Multi-view video generation with diffusion model.
\newblock \emph{Advances in Neural Information Processing Systems}, 37:\penalty0 62189--62222, 2025.

\bibitem[Li et~al.(2024{\natexlab{a}})Li, Tan, Zhang, Xu, Luan, Xu, Hong, Sunkavalli, Shakhnarovich, and Bi]{instant3d2023}
Jiahao Li, Hao Tan, Kai Zhang, Zexiang Xu, Fujun Luan, Yinghao Xu, Yicong Hong, Kalyan Sunkavalli, Greg Shakhnarovich, and Sai Bi.
\newblock Instant3d: Fast text-to-3d with sparse-view generation and large reconstruction model.
\newblock In \emph{The Twelfth International Conference on Learning Representations}, 2024{\natexlab{a}}.

\bibitem[Li et~al.(2024{\natexlab{b}})Li, Chen, Chen, and Tan]{li2023sweetdreamer}
Weiyu Li, Rui Chen, Xuelin Chen, and Ping Tan.
\newblock Sweetdreamer: Aligning geometric priors in 2d diffusion for consistent text-to-3d.
\newblock In \emph{The Twelfth International Conference on Learning Representations}, 2024{\natexlab{b}}.

\bibitem[Li et~al.(2023)Li, Zhu, Han, Hou, Guo, and Cheng]{licvpr23amt}
Zhen Li, Zuo-Liang Zhu, Ling-Hao Han, Qibin Hou, Chun-Le Guo, and Ming-Ming Cheng.
\newblock Amt: All-pairs multi-field transforms for efficient frame interpolation.
\newblock In \emph{IEEE Conference on Computer Vision and Pattern Recognition}, 2023.

\bibitem[Liang et~al.(2024{\natexlab{a}})Liang, Yin, Xu, hanxue liang, Wang, Plataniotis, Zhao, and Wei]{liang2024diffusion4d}
Hanwen Liang, Yuyang Yin, Dejia Xu, hanxue liang, Zhangyang Wang, Konstantinos~N Plataniotis, Yao Zhao, and Yunchao Wei.
\newblock Diffusion4d: Fast spatial-temporal consistent 4d generation via video diffusion models.
\newblock In \emph{The Thirty-eighth Annual Conference on Neural Information Processing Systems}, 2024{\natexlab{a}}.

\bibitem[Liang et~al.(2024{\natexlab{b}})Liang, Yang, Lin, Li, Xu, and Chen]{EnVision2023luciddreamer}
Yixun Liang, Xin Yang, Jiantao Lin, Haodong Li, Xiaogang Xu, and Yingcong Chen.
\newblock Luciddreamer: Towards high-fidelity text-to-3d generation via interval score matching.
\newblock In \emph{Proceedings of the IEEE/CVF conference on computer vision and pattern recognition}, pp.\  6517--6526, 2024{\natexlab{b}}.

\bibitem[Ling et~al.(2024)Ling, Kim, Torralba, Fidler, and Kreis]{ling2023alignyourgaussians}
Huan Ling, Seung~Wook Kim, Antonio Torralba, Sanja Fidler, and Karsten Kreis.
\newblock Align your gaussians: Text-to-4d with dynamic 3d gaussians and composed diffusion models.
\newblock In \emph{Proceedings of the IEEE/CVF conference on computer vision and pattern recognition}, pp.\  8576--8588, 2024.

\bibitem[Liu et~al.(2023{\natexlab{a}})Liu, Xu, Jin, Chen, Varma~T, Xu, and Su]{liu2024one}
Minghua Liu, Chao Xu, Haian Jin, Linghao Chen, Mukund Varma~T, Zexiang Xu, and Hao Su.
\newblock One-2-3-45: Any single image to 3{D} mesh in 45 seconds without per-shape optimization.
\newblock In \emph{Advances in Neural Information Processing Systems}, 2023{\natexlab{a}}.

\bibitem[Liu et~al.(2024{\natexlab{a}})Liu, Shi, Chen, Zhang, Xu, Wei, Chen, Zeng, Gu, and Su]{liu2023one2}
Minghua Liu, Ruoxi Shi, Linghao Chen, Zhuoyang Zhang, Chao Xu, Xinyue Wei, Hansheng Chen, Chong Zeng, Jiayuan Gu, and Hao Su.
\newblock One-2-3-45++: Fast single image to 3d objects with consistent multi-view generation and 3d diffusion.
\newblock In \emph{Proceedings of the IEEE/CVF conference on computer vision and pattern recognition}, pp.\  10072--10083, 2024{\natexlab{a}}.

\bibitem[Liu et~al.(2023{\natexlab{b}})Liu, Wu, Van~Hoorick, Tokmakov, Zakharov, and Vondrick]{liu2023zero}
Ruoshi Liu, Rundi Wu, Basile Van~Hoorick, Pavel Tokmakov, Sergey Zakharov, and Carl Vondrick.
\newblock Zero-1-to-3: Zero-shot one image to 3{D} object.
\newblock In \emph{Proceedings of the IEEE/CVF International Conference on Computer Vision}, 2023{\natexlab{b}}.

\bibitem[Liu et~al.(2024{\natexlab{b}})Liu, Lin, Zeng, Long, Liu, Komura, and Wang]{liu2023syncdreamer}
Yuan Liu, Cheng Lin, Zijiao Zeng, Xiaoxiao Long, Lingjie Liu, Taku Komura, and Wenping Wang.
\newblock Syncdreamer: Generating multiview-consistent images from a single-view image.
\newblock In \emph{The Twelfth International Conference on Learning Representations}, 2024{\natexlab{b}}.

\bibitem[Long et~al.(2024)Long, Guo, Lin, Liu, Dou, Liu, Ma, Zhang, Habermann, Theobalt, et~al.]{long2023wonder3d}
Xiaoxiao Long, Yuan-Chen Guo, Cheng Lin, Yuan Liu, Zhiyang Dou, Lingjie Liu, Yuexin Ma, Song-Hai Zhang, Marc Habermann, Christian Theobalt, et~al.
\newblock Wonder3d: Single image to 3d using cross-domain diffusion.
\newblock In \emph{Proceedings of the IEEE/CVF conference on computer vision and pattern recognition}, pp.\  9970--9980, 2024.

\bibitem[Melas-Kyriazi et~al.(2024)Melas-Kyriazi, Laina, Rupprecht, Neverova, Vedaldi, Gafni, and Kokkinos]{melas20243d}
Luke Melas-Kyriazi, Iro Laina, Christian Rupprecht, Natalia Neverova, Andrea Vedaldi, Oran Gafni, and Filippos Kokkinos.
\newblock Im-3d: Iterative multiview diffusion and reconstruction for high-quality 3d generation.
\newblock \emph{International Conference on Machine Learning, 2024}, 2024.

\bibitem[Mildenhall et~al.(2021)Mildenhall, Srinivasan, Tancik, Barron, Ramamoorthi, and Ng]{mildenhall2021nerf}
Ben Mildenhall, Pratul~P Srinivasan, Matthew Tancik, Jonathan~T Barron, Ravi Ramamoorthi, and Ren Ng.
\newblock {NeRF}: Representing scenes as neural radiance fields for view synthesis.
\newblock \emph{Communications of the ACM}, 2021.

\bibitem[M\"uller et~al.(2022)M\"uller, Evans, Schied, and Keller]{mueller2022instant}
Thomas M\"uller, Alex Evans, Christoph Schied, and Alexander Keller.
\newblock Instant neural graphics primitives with a multiresolution hash encoding.
\newblock \emph{ACM Trans. Graph.}, 2022.

\bibitem[Niemeyer et~al.(2022)Niemeyer, Barron, Mildenhall, Sajjadi, Geiger, and Radwan]{Niemeyer2021Regnerf}
Michael Niemeyer, Jonathan~T. Barron, Ben Mildenhall, Mehdi S.~M. Sajjadi, Andreas Geiger, and Noha Radwan.
\newblock {RegNeRF}: Regularizing neural radiance fields for view synthesis from sparse inputs.
\newblock In \emph{Proc. IEEE Conf. on Computer Vision and Pattern Recognition (CVPR)}, 2022.

\bibitem[Pan et~al.(2024{\natexlab{a}})Pan, Lu, Zhu, and Zhang]{pan2023enhancing}
Zijie Pan, Jiachen Lu, Xiatian Zhu, and Li~Zhang.
\newblock Enhancing high-resolution 3d generation through pixel-wise gradient clipping.
\newblock In \emph{The Twelfth International Conference on Learning Representations}, 2024{\natexlab{a}}.

\bibitem[Pan et~al.(2024{\natexlab{b}})Pan, Yang, Zhu, and Zhang]{pan2024fast}
Zijie Pan, Zeyu Yang, Xiatian Zhu, and Li~Zhang.
\newblock Fast dynamic 3{D} object generation from a single-view video.
\newblock \emph{arXiv preprint arXiv:2401.08742}, 2024{\natexlab{b}}.

\bibitem[Perazzi et~al.(2016)Perazzi, Pont-Tuset, McWilliams, {Van Gool}, Gross, and Sorkine-Hornung]{Perazzi_CVPR_2016}
Federico Perazzi, Jordi Pont-Tuset, Brian McWilliams, Luc {Van Gool}, Markus Gross, and Alexander Sorkine-Hornung.
\newblock A benchmark dataset and evaluation methodology for video object segmentation.
\newblock In \emph{The IEEE Conference on Computer Vision and Pattern Recognition}, 2016.

\bibitem[Pont-Tuset et~al.(2017)Pont-Tuset, Perazzi, Caelles, Arbel\'aez, Sorkine-Hornung, and {Van Gool}]{Pont-Tuset_arXiv_2017}
Jordi Pont-Tuset, Federico Perazzi, Sergi Caelles, Pablo Arbel\'aez, Alexander Sorkine-Hornung, and Luc {Van Gool}.
\newblock The 2017 davis challenge on video object segmentation.
\newblock \emph{arXiv:1704.00675}, 2017.

\bibitem[Poole et~al.(2023)Poole, Jain, Barron, and Mildenhall]{poole2022dreamfusion}
Ben Poole, Ajay Jain, Jonathan~T Barron, and Ben Mildenhall.
\newblock Dreamfusion: Text-to-3d using 2d diffusion.
\newblock In \emph{The Eleventh International Conference on Learning Representations}, 2023.

\bibitem[Pumarola et~al.(2021)Pumarola, Corona, Pons-Moll, and Moreno-Noguer]{pumarola2021d}
Albert Pumarola, Enric Corona, Gerard Pons-Moll, and Francesc Moreno-Noguer.
\newblock {D-NeRF}: Neural radiance fields for dynamic scenes.
\newblock In \emph{Proceedings of the IEEE/CVF Conference on Computer Vision and Pattern Recognition}, 2021.

\bibitem[Radford et~al.(2021)Radford, Kim, Hallacy, Ramesh, Goh, Agarwal, Sastry, Askell, Mishkin, Clark, et~al.]{radford2021learning}
Alec Radford, Jong~Wook Kim, Chris Hallacy, Aditya Ramesh, Gabriel Goh, Sandhini Agarwal, Girish Sastry, Amanda Askell, Pamela Mishkin, Jack Clark, et~al.
\newblock Learning transferable visual models from natural language supervision.
\newblock In \emph{International conference on machine learning}, 2021.

\bibitem[Ren et~al.(2023)Ren, Pan, Tang, Zhang, Cao, Zeng, and Liu]{ren2023dreamgaussian4d}
Jiawei Ren, Liang Pan, Jiaxiang Tang, Chi Zhang, Ang Cao, Gang Zeng, and Ziwei Liu.
\newblock {DreamGaussian4D}: Generative 4{D} gaussian splatting.
\newblock \emph{arXiv preprint arXiv:2312.17142}, 2023.

\bibitem[Ren et~al.(2025)Ren, Xie, Mirzaei, Kreis, Liu, Torralba, Fidler, Kim, Ling, et~al.]{ren2024l4gm}
Jiawei Ren, Cheng Xie, Ashkan Mirzaei, Karsten Kreis, Ziwei Liu, Antonio Torralba, Sanja Fidler, Seung~Wook Kim, Huan Ling, et~al.
\newblock L4gm: Large 4d gaussian reconstruction model.
\newblock \emph{Advances in Neural Information Processing Systems}, 37:\penalty0 56828--56858, 2025.

\bibitem[Rombach et~al.(2022)Rombach, Blattmann, Lorenz, Esser, and Ommer]{rombach2022high}
Robin Rombach, Andreas Blattmann, Dominik Lorenz, Patrick Esser, and Bj{\"o}rn Ommer.
\newblock High-resolution image synthesis with latent diffusion models.
\newblock In \emph{Proceedings of the IEEE/CVF conference on computer vision and pattern recognition}, 2022.

\bibitem[Sargent et~al.(2024)Sargent, Li, Shah, Herrmann, Yu, Zhang, Chan, Lagun, Fei-Fei, Sun, et~al.]{sargent2023zeronvs}
Kyle Sargent, Zizhang Li, Tanmay Shah, Charles Herrmann, Hong-Xing Yu, Yunzhi Zhang, Eric~Ryan Chan, Dmitry Lagun, Li~Fei-Fei, Deqing Sun, et~al.
\newblock Zeronvs: Zero-shot 360-degree view synthesis from a single image.
\newblock In \emph{Proceedings of the IEEE/CVF Conference on Computer Vision and Pattern Recognition}, pp.\  9420--9429, 2024.

\bibitem[Sauer et~al.(2025)Sauer, Lorenz, Blattmann, and Rombach]{sauer2025adversarial}
Axel Sauer, Dominik Lorenz, Andreas Blattmann, and Robin Rombach.
\newblock Adversarial diffusion distillation.
\newblock In \emph{European Conference on Computer Vision}, pp.\  87--103. Springer, 2025.

\bibitem[Shi et~al.(2023)Shi, Chen, Zhang, Liu, Xu, Wei, Chen, Zeng, and Su]{shi2023zero123++}
Ruoxi Shi, Hansheng Chen, Zhuoyang Zhang, Minghua Liu, Chao Xu, Xinyue Wei, Linghao Chen, Chong Zeng, and Hao Su.
\newblock Zero123++: a single image to consistent multi-view diffusion base model.
\newblock \emph{arXiv preprint arXiv:2310.15110}, 2023.

\bibitem[Shi et~al.(2024{\natexlab{a}})Shi, Wang, Ye, Mai, Li, and Yang]{shi2023mvdream}
Yichun Shi, Peng Wang, Jianglong Ye, Long Mai, Kejie Li, and Xiao Yang.
\newblock Mvdream: Multi-view diffusion for 3d generation.
\newblock In \emph{The Twelfth International Conference on Learning Representations}, 2024{\natexlab{a}}.

\bibitem[Shi et~al.(2024{\natexlab{b}})Shi, Wang, He, Tang, Qi, Yang, Huang, Liu, Zhang, and Shum]{shi2023toss}
Yukai Shi, Jianan Wang, CAO He, Boshi Tang, Xianbiao Qi, Tianyu Yang, Yukun Huang, Shilong Liu, Lei Zhang, and Heung-Yeung Shum.
\newblock Toss: High-quality text-guided novel view synthesis from a single image.
\newblock In \emph{The Twelfth International Conference on Learning Representations}, 2024{\natexlab{b}}.

\bibitem[Singer et~al.(2023{\natexlab{a}})Singer, Polyak, Hayes, Yin, An, Zhang, Hu, Yang, Ashual, Gafni, et~al.]{singer2022make}
Uriel Singer, Adam Polyak, Thomas Hayes, Xi~Yin, Jie An, Songyang Zhang, Qiyuan Hu, Harry Yang, Oron Ashual, Oran Gafni, et~al.
\newblock Make-a-video: Text-to-video generation without text-video data.
\newblock In \emph{The Eleventh International Conference on Learning Representations}, 2023{\natexlab{a}}.

\bibitem[Singer et~al.(2023{\natexlab{b}})Singer, Sheynin, Polyak, Ashual, Makarov, Kokkinos, Goyal, Vedaldi, Parikh, Johnson, et~al.]{singer2023text4d}
Uriel Singer, Shelly Sheynin, Adam Polyak, Oron Ashual, Iurii Makarov, Filippos Kokkinos, Naman Goyal, Andrea Vedaldi, Devi Parikh, Justin Johnson, et~al.
\newblock Text-to-4d dynamic scene generation.
\newblock In \emph{Proceedings of the 40th International Conference on Machine Learning}, pp.\  31915--31929, 2023{\natexlab{b}}.

\bibitem[Song et~al.(2021)Song, Meng, and Ermon]{song2020denoising}
Jiaming Song, Chenlin Meng, and Stefano Ermon.
\newblock Denoising diffusion implicit models.
\newblock In \emph{International Conference on Learning Representations}, 2021.

\bibitem[Sun et~al.(2024)Sun, Zhang, Shao, Wang, Liu, Xie, and Liu]{sun2023dreamcraft3d}
Jingxiang Sun, Bo~Zhang, Ruizhi Shao, Lizhen Wang, Wen Liu, Zhenda Xie, and Yebin Liu.
\newblock Dreamcraft3d: Hierarchical 3d generation with bootstrapped diffusion prior.
\newblock In \emph{The Twelfth International Conference on Learning Representations}, 2024.

\bibitem[Sun et~al.(2025)Sun, Guo, Wan, Yan, Yin, Zhou, Liao, and Li]{sun2024eg4d}
Qi~Sun, Zhiyang Guo, Ziyu Wan, Jing~Nathan Yan, Shengming Yin, Wengang Zhou, Jing Liao, and Houqiang Li.
\newblock {EG}4d: Explicit generation of 4d object without score distillation.
\newblock In \emph{The Thirteenth International Conference on Learning Representations}, 2025.

\bibitem[Tang et~al.(2024)Tang, Ren, Zhou, Liu, and Zeng]{tang2023dreamgaussian}
Jiaxiang Tang, Jiawei Ren, Hang Zhou, Ziwei Liu, and Gang Zeng.
\newblock Dreamgaussian: Generative gaussian splatting for efficient 3d content creation.
\newblock In \emph{The Twelfth International Conference on Learning Representations}, 2024.

\bibitem[Tochilkin et~al.(2024)Tochilkin, Pankratz, Liu, Huang, Letts, Li, Liang, Laforte, Jampani, and Cao]{tochilkin2024triposr}
Dmitry Tochilkin, David Pankratz, Zexiang Liu, Zixuan Huang, Adam Letts, Yangguang Li, Ding Liang, Christian Laforte, Varun Jampani, and Yan-Pei Cao.
\newblock {TripoSR}: Fast 3{D} object reconstruction from a single image.
\newblock \emph{arXiv preprint arXiv:2403.02151}, 2024.

\bibitem[Unterthiner et~al.(2018)Unterthiner, Van~Steenkiste, Kurach, Marinier, Michalski, and Gelly]{unterthiner2018towards}
Thomas Unterthiner, Sjoerd Van~Steenkiste, Karol Kurach, Raphael Marinier, Marcin Michalski, and Sylvain Gelly.
\newblock Towards accurate generative models of video: A new metric \& challenges.
\newblock \emph{arXiv preprint arXiv:1812.01717}, 2018.

\bibitem[Voleti et~al.(2022)Voleti, Jolicoeur-Martineau, and Pal]{voleti2022mcvd}
Vikram Voleti, Alexia Jolicoeur-Martineau, and Christopher Pal.
\newblock {MCVD}: Masked conditional video diffusion for prediction, generation, and interpolation.
\newblock In \emph{Advances in Neural Information Processing Systems}, 2022.

\bibitem[Voleti et~al.(2024)Voleti, Yao, Boss, Letts, Pankratz, Tochilkin, Laforte, Rombach, and Jampani]{voleti2024sv3d}
Vikram Voleti, Chun-Han Yao, Mark Boss, Adam Letts, David Pankratz, Dmitrii Tochilkin, Christian Laforte, Robin Rombach, and Varun Jampani.
\newblock {SV3D}: Novel multi-view synthesis and {3D} generation from a single image using latent video diffusion.
\newblock In \emph{European Conference on Computer Vision}, 2024.

\bibitem[Wang \& Shi(2023)Wang and Shi]{wang2023imagedream}
Peng Wang and Yichun Shi.
\newblock {ImageDream}: Image-prompt multi-view diffusion for 3{D} generation.
\newblock \emph{arXiv preprint arXiv:2312.02201}, 2023.

\bibitem[Wang et~al.(2024{\natexlab{a}})Wang, Tan, Bi, Xu, Luan, Sunkavalli, Wang, Xu, and Zhang]{wang2023pf}
Peng Wang, Hao Tan, Sai Bi, Yinghao Xu, Fujun Luan, Kalyan Sunkavalli, Wenping Wang, Zexiang Xu, and Kai Zhang.
\newblock Pf-lrm: Pose-free large reconstruction model for joint pose and shape prediction.
\newblock In \emph{The Twelfth International Conference on Learning Representations}, 2024{\natexlab{a}}.

\bibitem[Wang et~al.(2025)Wang, Wang, Chen, Wang, Sun, and Zhu]{wang2024vidu4d}
Yikai Wang, Xinzhou Wang, Zilong Chen, Zhengyi Wang, Fuchun Sun, and Jun Zhu.
\newblock Vidu4d: Single generated video to high-fidelity 4d reconstruction with dynamic gaussian surfels.
\newblock \emph{Advances in Neural Information Processing Systems}, 37:\penalty0 131316--131343, 2025.

\bibitem[Wang et~al.(2024{\natexlab{b}})Wang, Lu, Wang, Bao, Li, Su, and Zhu]{wang2024prolificdreamer}
Zhengyi Wang, Cheng Lu, Yikai Wang, Fan Bao, Chongxuan Li, Hang Su, and Jun Zhu.
\newblock {ProlificDreamer}: High-fidelity and diverse text-to-3{D} generation with variational score distillation.
\newblock In \emph{Advances in Neural Information Processing Systems}, 2024{\natexlab{b}}.

\bibitem[Wei et~al.(2024)Wei, Zhang, Bi, Tan, Luan, Deschaintre, Sunkavalli, Su, and Xu]{wei2024meshlrm}
Xinyue Wei, Kai Zhang, Sai Bi, Hao Tan, Fujun Luan, Valentin Deschaintre, Kalyan Sunkavalli, Hao Su, and Zexiang Xu.
\newblock {MeshLRM}: Large reconstruction model for high-quality mesh.
\newblock \emph{arXiv preprint arXiv:2404.12385}, 2024.

\bibitem[Weng et~al.(2023)Weng, Yang, Wang, Li, Zhang, Chen, and Zhang]{weng2023consistent123}
Haohan Weng, Tianyu Yang, Jianan Wang, Yu~Li, Tong Zhang, CL~Chen, and Lei Zhang.
\newblock Consistent123: Improve consistency for one image to 3{D} object synthesis.
\newblock \emph{arXiv preprint arXiv:2310.08092}, 2023.

\bibitem[Wu et~al.(2024)Wu, Yi, Fang, Xie, Zhang, Wei, Liu, Tian, and Wang]{wu20244d}
Guanjun Wu, Taoran Yi, Jiemin Fang, Lingxi Xie, Xiaopeng Zhang, Wei Wei, Wenyu Liu, Qi~Tian, and Xinggang Wang.
\newblock {4D} gaussian splatting for real-time dynamic scene rendering.
\newblock In \emph{Proceedings of the IEEE/CVF Conference on Computer Vision and Pattern Recognition}, 2024.

\bibitem[Yang et~al.(2025)Yang, Pan, Gu, and Zhang]{yang2024diffusion}
Zeyu Yang, Zijie Pan, Chun Gu, and Li~Zhang.
\newblock Diffusion\${\textasciicircum}2\$: Dynamic 3d content generation via score composition of video and multi-view diffusion models.
\newblock In \emph{The Thirteenth International Conference on Learning Representations}, 2025.

\bibitem[Ye et~al.(2024)Ye, Wang, Li, Shi, and Wang]{ye2023consistent}
Jianglong Ye, Peng Wang, Kejie Li, Yichun Shi, and Heng Wang.
\newblock Consistent-1-to-3: Consistent image to 3d view synthesis via geometry-aware diffusion models.
\newblock In \emph{2024 International Conference on 3D Vision (3DV)}, pp.\  664--674. IEEE, 2024.

\bibitem[Yi et~al.(2024)Yi, Fang, Wang, Wu, Xie, Zhang, Liu, Tian, and Wang]{yi2023gaussiandreamer}
Taoran Yi, Jiemin Fang, Junjie Wang, Guanjun Wu, Lingxi Xie, Xiaopeng Zhang, Wenyu Liu, Qi~Tian, and Xinggang Wang.
\newblock Gaussiandreamer: Fast generation from text to 3d gaussians by bridging 2d and 3d diffusion models.
\newblock In \emph{Proceedings of the IEEE/CVF Conference on Computer Vision and Pattern Recognition}, pp.\  6796--6807, 2024.

\bibitem[Yin et~al.(2023)Yin, Xu, Wang, Zhao, and Wei]{yin20234dgen}
Yuyang Yin, Dejia Xu, Zhangyang Wang, Yao Zhao, and Yunchao Wei.
\newblock {4DGen}: Grounded 4{D} content generation with spatial-temporal consistency.
\newblock \emph{arXiv preprint arXiv:2312.17225}, 2023.

\bibitem[Yu et~al.(2022)Yu, Peng, Niemeyer, Sattler, and Geiger]{yu2022monosdf}
Zehao Yu, Songyou Peng, Michael Niemeyer, Torsten Sattler, and Andreas Geiger.
\newblock {MonoSDF}: Exploring monocular geometric cues for neural implicit surface reconstruction.
\newblock \emph{Advances in neural information processing systems}, 2022.

\bibitem[Zeng et~al.(2024)Zeng, Jiang, Zhu, Lu, Lin, Zhu, Hu, Cao, and Yao]{zeng2024stag4d}
Yifei Zeng, Yanqin Jiang, Siyu Zhu, Yuanxun Lu, Youtian Lin, Hao Zhu, Weiming Hu, Xun Cao, and Yao Yao.
\newblock Stag4d: Spatial-temporal anchored generative 4d gaussians.
\newblock In \emph{European Conference on Computer Vision}, pp.\  163--179. Springer, 2024.

\bibitem[Zhang et~al.(2024)Zhang, Chen, Wang, Liu, Wang, and Qiao]{zhang20244diffusion}
Haiyu Zhang, Xinyuan Chen, Yaohui Wang, Xihui Liu, Yunhong Wang, and Yu~Qiao.
\newblock 4diffusion: Multi-view video diffusion model for 4d generation.
\newblock \emph{Advances in Neural Information Processing Systems}, 37:\penalty0 15272--15295, 2024.

\bibitem[Zhang et~al.(2018{\natexlab{a}})Zhang, Isola, Efros, Shechtman, and Wang]{zhang2018lpips}
Richard Zhang, Phillip Isola, Alexei~A Efros, Eli Shechtman, and Oliver Wang.
\newblock The unreasonable effectiveness of deep features as a perceptual metric.
\newblock In \emph{Proceedings of the IEEE conference on computer vision and pattern recognition}, pp.\  586--595, 2018{\natexlab{a}}.

\bibitem[Zhang et~al.(2018{\natexlab{b}})Zhang, Isola, Efros, Shechtman, and Wang]{zhang2018unreasonable}
Richard Zhang, Phillip Isola, Alexei~A Efros, Eli Shechtman, and Oliver Wang.
\newblock The unreasonable effectiveness of deep features as a perceptual metric.
\newblock In \emph{Proceedings of the IEEE conference on computer vision and pattern recognition}, 2018{\natexlab{b}}.

\bibitem[Zhao et~al.(2023)Zhao, Yan, Xie, Hong, Li, and Lee]{zhao2023animate124}
Yuyang Zhao, Zhiwen Yan, Enze Xie, Lanqing Hong, Zhenguo Li, and Gim~Hee Lee.
\newblock Animate124: Animating one image to 4{D} dynamic scene.
\newblock \emph{arXiv preprint arXiv:2311.14603}, 2023.

\bibitem[Zhou et~al.(2024)Zhou, Shih, Meng, and Ermon]{zhou2023dreampropeller}
Linqi Zhou, Andy Shih, Chenlin Meng, and Stefano Ermon.
\newblock Dreampropeller: Supercharge text-to-3d generation with parallel sampling.
\newblock In \emph{Proceedings of the IEEE/CVF Conference on Computer Vision and Pattern Recognition}, pp.\  4610--4619, 2024.

\bibitem[Zou et~al.(2024)Zou, Yu, Guo, Li, Liang, Cao, and Zhang]{zou2023triplane}
Zi-Xin Zou, Zhipeng Yu, Yuan-Chen Guo, Yangguang Li, Ding Liang, Yan-Pei Cao, and Song-Hai Zhang.
\newblock Triplane meets gaussian splatting: Fast and generalizable single-view 3d reconstruction with transformers.
\newblock In \emph{Proceedings of the IEEE/CVF conference on computer vision and pattern recognition}, pp.\  10324--10335, 2024.

\end{thebibliography}
\bibliographystyle{iclr2025_conference}

\newpage
\appendix

\section{Implementation Details}

\subsection{Data Details}
\label{supp:data_details}
Similar to SV3D~\citep{voleti2024sv3d}, we render views of a subset of the Objaverse dataset~\citep{deitke2023objaverse} curated CC-licensed animatable 3D objects from the original dataset. 
We held out some 
objects for evaluation purposes.

Each loaded object is scaled such the largest world-space XYZ extent of its bounding box is 1.
We use Blender's CYCLES renderer to render 
multiple views and 
video frames for each object.
We limit the number of samples in CYCLES renderer to save the rendering time.
For lightning, we follow SV3D to randomly select from a set of 
curated HDRI envmaps. 

For static orbits, we regularly sample azimuth angles with a constant value
starting from azimuth 0. 
We randomly sample one elevation angle 
for all 
views. 
For dynamic orbits, the sequence of camera elevations for each orbit is obtained from a random weighted combination of sinusoids with different frequencies~\citep{voleti2024sv3d}. 
The azimuth angles are sampled regularly, and then a small amount of noise is added to make them irregular. 
The elevation values are smoothed using a simple convolution kernel and then clamped to a maximum elevation. 

We then encode all of these images into latent space using SD2.1~\citep{rombach2022high}'s VAE, and CLIP~\citep{radford2021learning}.
We then store the latent and CLIP embeddings for all of these images along with the corresponding elevation and azimuth values, and frame index.

\subsection{SV4D Training Details}
\label{supp:training_details}
Our approach involves utilizing the widely used EMD~\citep{karras2022elucidating}, incorporating a $L2$ loss for fine-tuning, as followed in SVD~\citep{blattmann2023align} and SV3D~\citep{voleti2024sv3d}.
We adopt $L2$ loss during the training. 
To optimize the training efficiency and reduce GPU VRAM, we follow SV3D to reprocess and store the precomputed latent and CLIP embeddings for all images in advance.
During training, these tensors are directly loaded rather than being computed in real-time.
We found that with 8 views and 5 frames we were able to fit a batch size of 1 at $576\times576$ resolution. 
We randomly sample 8 views and 5 frames from our 21 rendered views and 21 frames for training.

\subsection{SV4D Inference Details}
\label{supp:inference_details}
During inference, we use 50 steps of the deterministic DDIM sampler~\citep{song2020denoising}. 
We adopt the SV4D sampling to generate the full V$\times$F image matrix during the inference.
We take $V=8$ and $F=21$ as an example. 
Our SV4D can generate 8 views $\times$ 5 frames in an inference.
In the interleaved sampling, we first generate a sparse set of anchor frames with the frame indexes [4, 8, 12, 16, 20].
Then in the dense sampling, we use the anchor frames as new reference images to densely sample the remaining frames.
For example, in the first run, we generate frames with the indexes [0, 1, 2, 3, 4] with frame 0 (forward) or frame 4 (backward) as conditioning.
It takes about 40 seconds to generate 5 frames and 8 views for the SV4D model.

\subsection{SV4D Network Architecture}
\label{supp:network_details}
SV4D network consists of three attention layers: spatial attention, view attention, and frame attention.
We reshape the feature $\mathbf{F}$ before feeding it to each attention layer $\gamma_{(\cdot)} (\cdot)$. 

The spatial attention models spatial coherence across image spatial locations by calculating the attention of points at the same image: 
\begin{align}
&  \mathbf{F} = \text{rearrange}(\mathbf{F}, ~ B~V~F~H~W~D \rightarrow  (B~V~F)~(H~W)~D)\\
&  \mathbf{F} =  \gamma_s (\mathbf{F}) \\
&  \mathbf{F} = \text{rearrange}(\mathbf{F}, ~(B~V~F)~(H~W)~D \rightarrow ~ B~V~F~H~W~D )
\end{align}

The view attention models coherence across views by calculating the attention of points at the same spatial location and the same frame across views:
\begin{align}
&  \mathbf{F} = \text{rearrange}(\mathbf{F}, ~ B~V~F~H~W~D \rightarrow  (B~F~H~W)~V~D)\\
&  \mathbf{F} =  \gamma_v (\mathbf{F}) \\
&  \mathbf{F} = \text{rearrange}(\mathbf{F}, ~ (B~F~H~W)~V~D \rightarrow ~ B~V~F~H~W~D )
\end{align}

The frame attention models coherence across frames by calculating the attention of points at the same spatial location and the same view across frames:
\begin{align}
&  \mathbf{F} = \text{rearrange}(\mathbf{F}, ~ B~V~F~H~W~D \rightarrow  (B~V~H~W)~F~D)\\
&  \mathbf{F} =  \gamma_f (\mathbf{F}) \\
&  \mathbf{F} = \text{rearrange}(\mathbf{F}, ~ (B~V~H~W)~F~D \rightarrow ~ B~V~F~H~W~D )
\end{align}
$B$ is batch size, $V$ is number of views, $F$ is number of frames. $H$, $W$, and $D$ are width, height, and feature dimension of the image feature.
In our experiments, both $V$ and $F$ are 8 and 5, respectively.

\newadd{
Note that our frame attention layers are skip-connected to the outputs of view attention with a learnable blending weight per layer, allowing it to effectively merge the spatial and temporal information in the final output. 
Although an optimal way is to perform spatial-temporal attention jointly without considering memory limitation, we find that our sequential design can best leverage the priors in SVD and SV3D with minimal computation overhead. 
}

\begin{figure}[t!]
    \centering
    \includegraphics[width=0.85\linewidth]{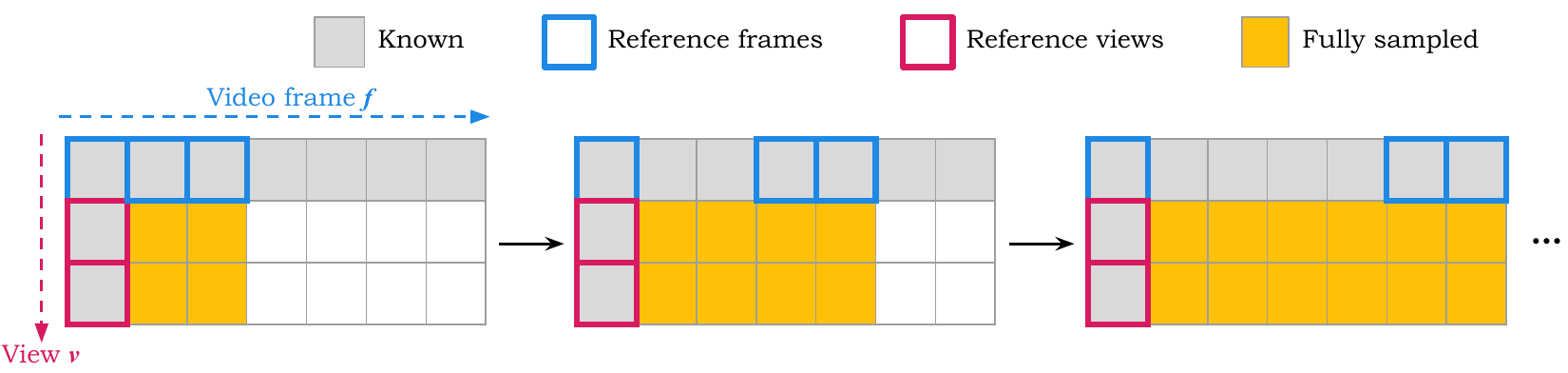}
    \caption{
        \textbf{Independent sampling}. The submatrix sampled at each inference consists of consecutive frames and each submatrix is sampled separately, without relying on previous fully sampled images.
    }
\label{fig:sampling_independent}
\end{figure}

\begin{figure}[t!]
    \centering
    \includegraphics[width=\linewidth]{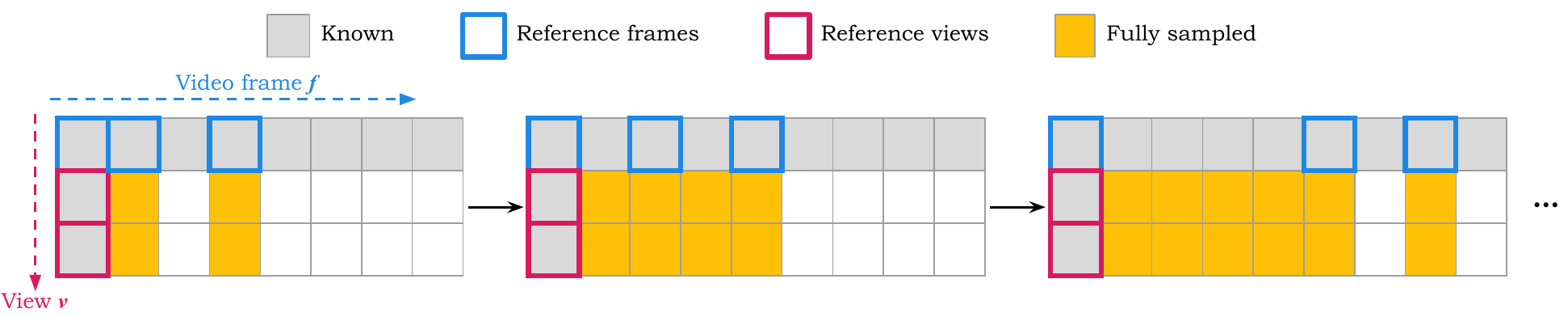}
    \caption{
        \textbf{Interleaved Sampling}. The submatrix sampled at each inference consists of interleaved frames and each submatrix is sampled separately, without relying on previous fully sampled images.
    }
\label{fig:sampling_interleaved}
\end{figure}

\subsection{Dynamic NeRF Optimization Details}
\label{supp:opt_details}
Our main 4D reconstruction losses are the pixel-level mean squared error $\mathcal{L}_\text{mse} = \lVert \bm{M}-\bm{\hat{M}} \rVert^2$, LPIPS~\citep{zhang2018lpips} loss $\mathcal{L}_\text{lpips}$, and mask loss $\mathcal{L}_\text{mask} = \lVert \bm{S} - \bm{\hat{S}} \rVert^2$, where $\bm{S}$, $\bm{\hat{S}}$ are the predicted and ground-truth masks.
We further employ a normal loss using the estimated mono normal by Omnidata~\citep{Eftekhar2021omnidata}, which is defined as the cosine similarity between the rendered normal $\bm{n}$ and estimated pseudo ground truths $\bm{\bar{n}}$: $\mathcal{L}_\text{normal}=1-(\bm{n} \cdot \bm{\bar{n}})$.
To regularize the output geometry, 
we apply a smooth depth loss inspired by RegNeRF~\citep{Niemeyer2021Regnerf}:
$\mathcal{L}_\text{depth}(i,j)=\left(d(i,j)-d(i,j+1)\right)^2 + \left(d(i,j)-(d(i+1,j)\right)^2$, where $i,j$ indicate the pixel coordinate.
For surface normal we instead rely on a bilateral smoothness loss similar to~\cite{boss2022-samurai}. We found that this is crucial to getting high-frequency details and avoiding over-smoothed surfaces. For this loss we compute the image gradients of the input image $\nabla \bm{I}$ with a Sobel filter~\citep{kanopoulos1988sobel}. We then encourage the gradients of rendered normal $\nabla \bm{n}$ to be smooth if (and only if) the input image gradients $\nabla \bm{I}$ are smooth. The loss can be written as $\mathcal{L}_\text{bilateral}=e^{-3 \nabla \bm{I}} \sqrt{1 + || \nabla \bm{n} ||}$.
The overall objective is then defined as the weighted sum of these losses. All losses are applied in both static and dynamic stages.
We use an Adam optimizer~\citep{kingma2014adam} with a learning rate of $0.01$ for both stages.

\subsection{Different Sampling Strategies}
\label{supp:sampling_details}
Due to memory limitations, we cannot generate all the novel view frames at once.
To generate the full image matrix, one can run the submatrix generation in different strategies.
To validate the effectiveness of our SV4D sampling, we compare ours with two other sampling strategies: independent sampling and interleaved sampling, which are shown in Fig.~\ref{fig:sampling_independent} and Fig.~\ref{fig:sampling_interleaved}, respectively.
For illustrative purposes only, the SV4D model can generate 2 frames and 2 views at each inference.
In the independent sampling, the submatrix sampled at each inference consists of \textit{consecutive} frames and each submatrix images are sampled separately, without relying on previous fully sampled images.
In the interleaved sampling, the submatrix sampled at each inference consists of \textit{interleaved} frames.

\section{Additional Results}
\begin{figure}[t!]
    \centering
    \includegraphics[width=\linewidth]{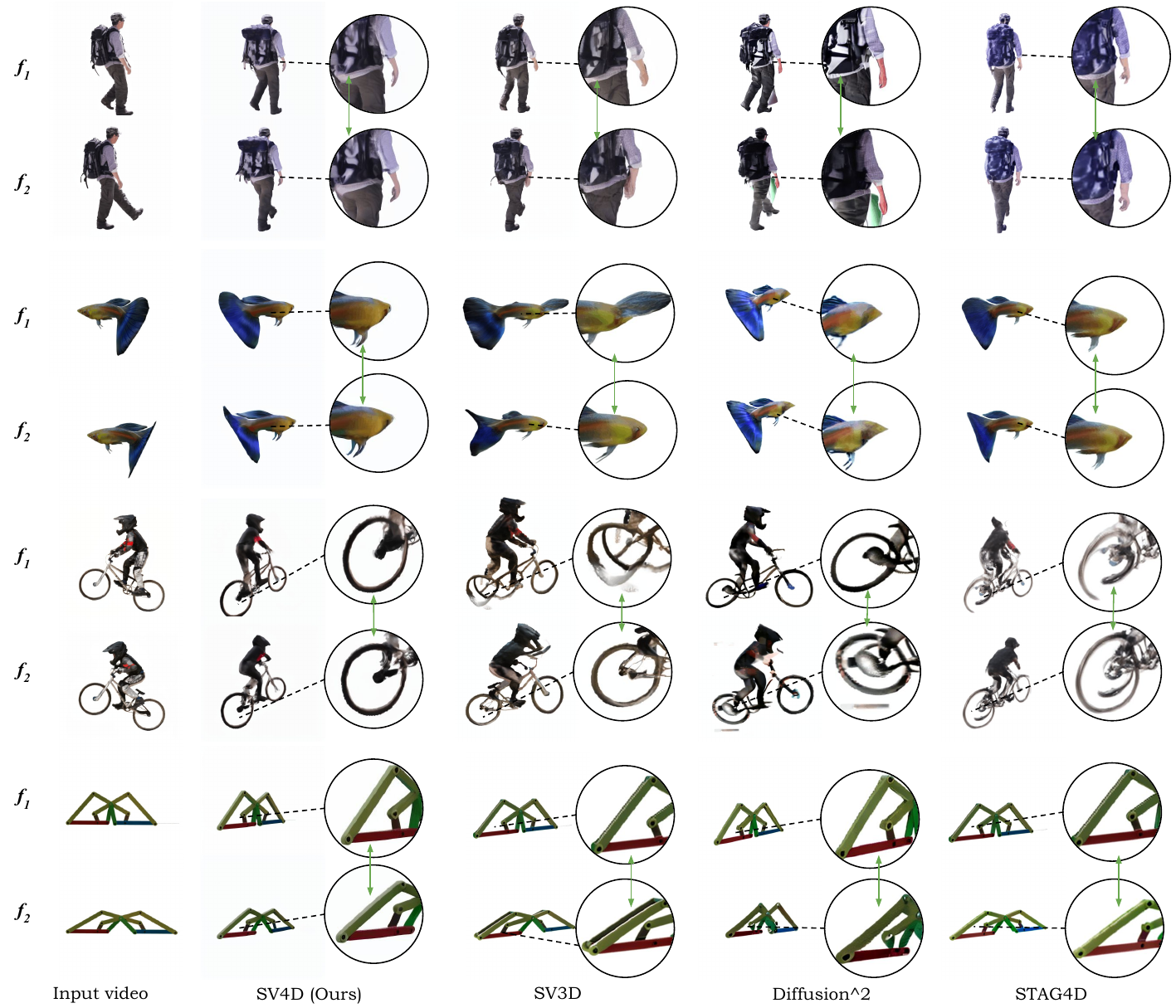}
    \vspace{-1.75em}
    \caption{
        \textbf{Visual Comparison of Novel View Video Synthesis Results.}
        We show two frames in the input videos and one novel-view result of the corresponding frames. Compared to the baseline methods, SV4D outputs contain geometry and texture details that are more faithful to the input video and consistent across frames.
    }
\label{fig:nvs_results_2}
\end{figure}
\begin{figure}[t!]
    \centering
    \includegraphics[width=\linewidth]{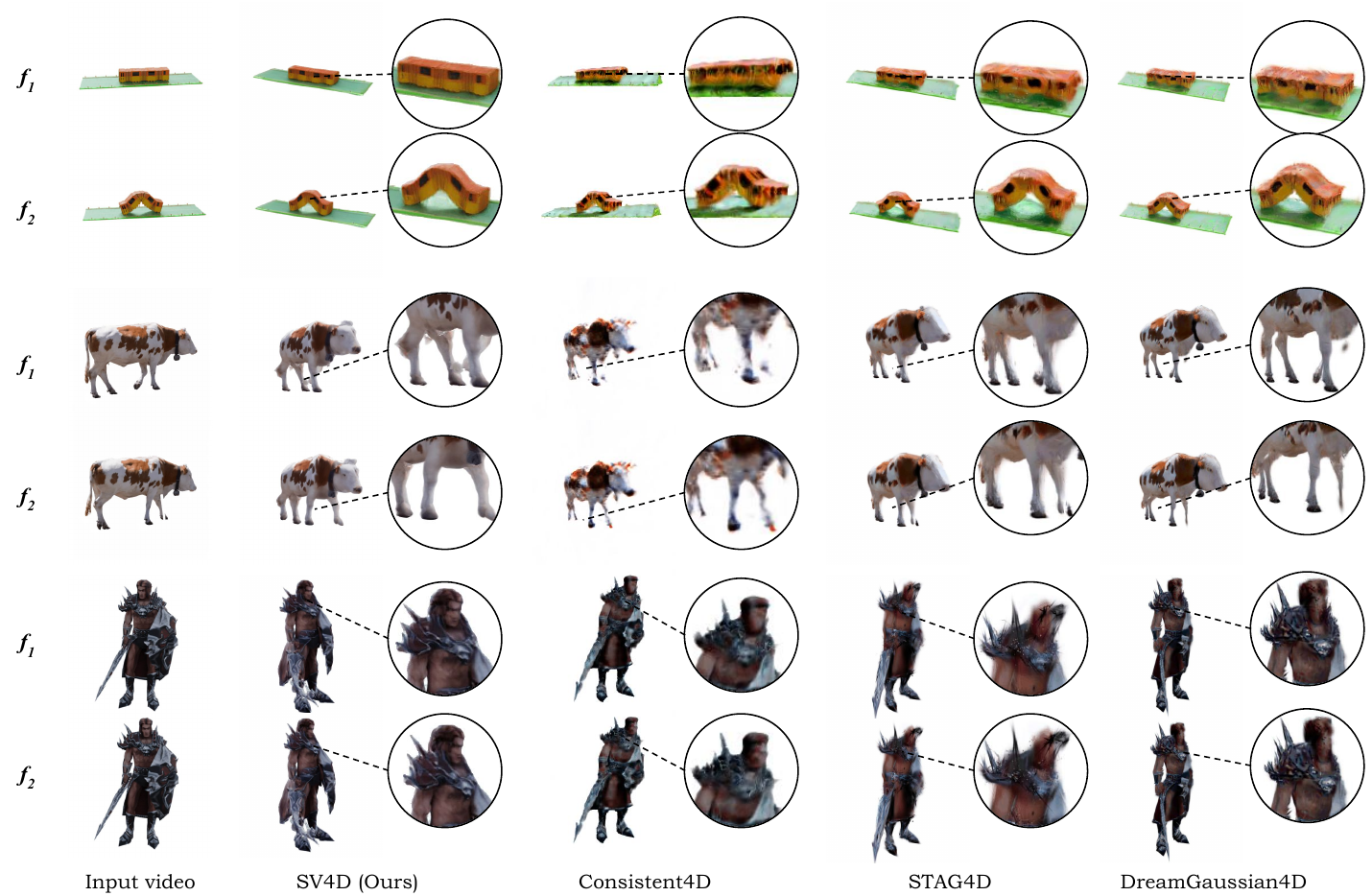}
    \vspace{-1.75em}
    \caption{
        \textbf{Visual Comparison of Generated 4D Outputs.} 
        We show two frames in the input videos and render one view result of the corresponding frames. 
        By leveraging the consistent multi-view videos generated by SV4D, we can learn higher-quality 4D assets compared to the prior works with SDS-based losses. Our results are more detailed, consistent, and faithful to the input videos.
    }
\label{fig:4d_results_2}
\end{figure}
\subsection{More Visual Comparisons}
\label{supp:additional_results}
We show more visual comparisons of novel view synthesis and 4D generation, as shown in Fig.~\ref{fig:nvs_results_2} and Fig.~\ref{fig:4d_results_2}.
These results further demonstrate that SV4D-generated images and 4D results are more consistent and detailed, faithful to the conditioning videos compared to the prior works.

\subsection{More Ablative Results}
\label{supp:additional_ablation}
\begin{table}[t!]
\caption{
\newadd{
\textbf{Additional Ablations.}
\textit{SV4D w/o frame atten.}: SV4D without frame attention.
\textit{SV4D w/o MV}: SV4D without reference multi-views.
\textit{SV4D with CLIP MV}: SV4D with CLIP embeddings (instead of VAE latents) of the first frame for frame attention conditioning and without reference multi-views.
}
}
\vspace{-.75em}
\label{tab:additional-ablation}
\centering
\resizebox{0.91\textwidth}{!}{
\begin{tabular}{ l c c c c c c c }
\toprule 
 Model & LPIPS$\downarrow$ & CLIP-S$\uparrow$ & FVD-F$\downarrow$ & FVD-V$\downarrow$ & FVD-Diag$\downarrow$ & FV4D$\downarrow$ \\
 \midrule
 SV4D w/o frame atten. & \textbf{0.131} & \textbf{0.920} & 729.67 & 375.49 & 526.78 & 690.49 \\
 SV4D w/o MV & 0.135 & 0.870 & 876.07 & 410.90 & 608.93 & 710.66 \\
 SV4D with CLIP MV & 0.142 & 0.868 & 1174.47 & 530.74 & 819.22 & 1327.75 \\
 SV4D & 0.136 & \textbf{0.920} & \textbf{585.09} & \textbf{331.94} & \textbf{503.02} & \textbf{470.46} \\
\bottomrule
\end{tabular}
}
\end{table}
\newadd{
We show an additional ablation study on the SV4D network architecture to justify our model design in \Cref{tab:additional-ablation}.
In particular, we report the quantitative comparison of ablated models: SV4D without frame attention, SV4D without multi-view conditioning, and SV4D with CLIP embeddings (instead of VAE latents) of the first frame for frame attention conditioning while removing the reference multi-views. The results justify that using VAE latents of reference multi-views as frame attention conditioning achieves better performance in most metrics compared to other design choices.
}

\begin{figure}[t!]
    \centering
    \includegraphics[width=0.5\linewidth]{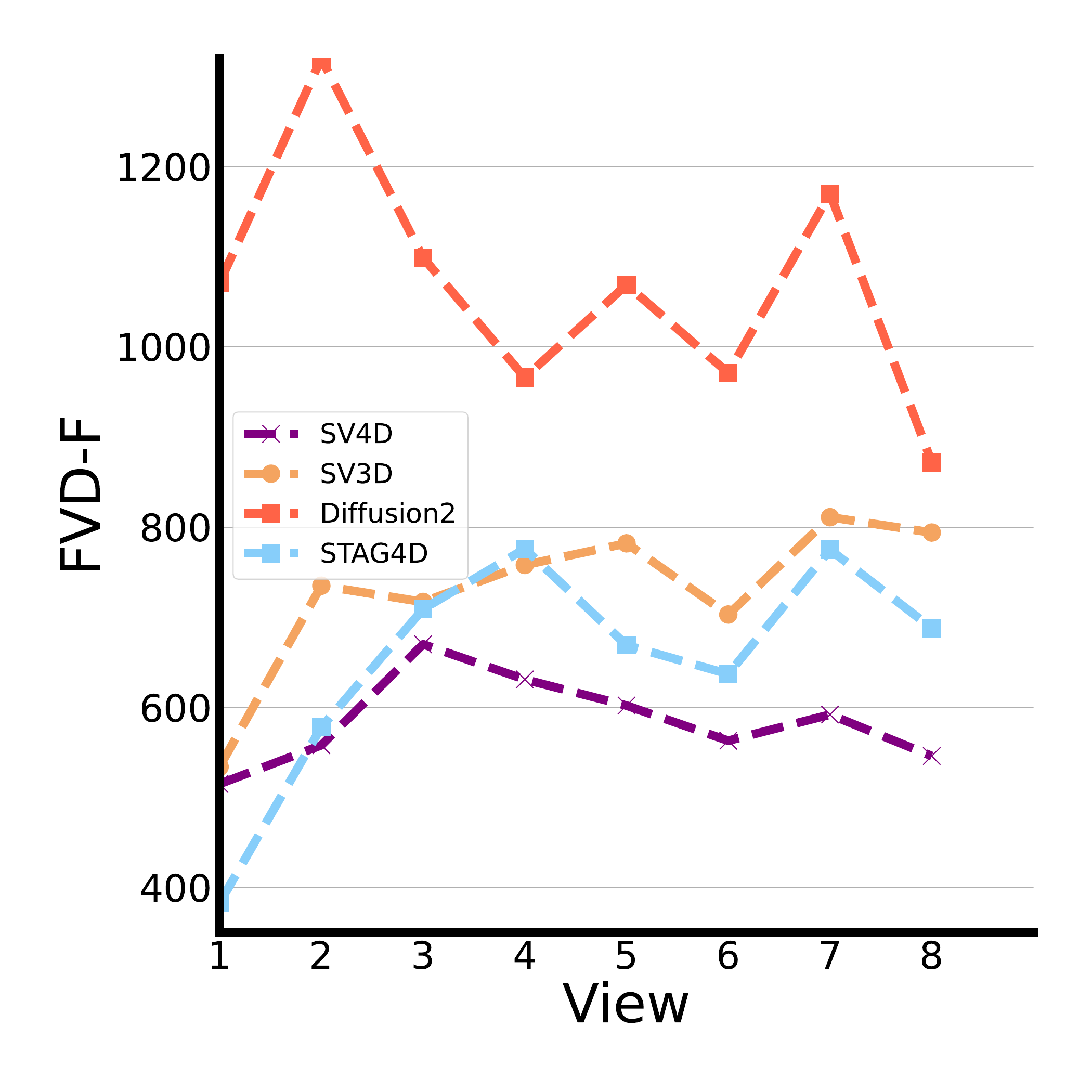}
    \vspace{-1.75em}
    \caption{
        \textbf{FVD-F vs. View Number} on the ObjaverseDy dataset. We observe that SV4D produces better temporal consistency in almost all views.
    }
\label{fig:fvd-view}
\end{figure}
\subsection{Quality Per View}
\label{supp:quality_per_view}
We plot the FVD-F value for each view on the ObjaverseDy dataset, as shown in Fig.~\ref{fig:fvd-view}.
We observe that SV4D produces better temporal consistency in almost all views.

\section{Baseline Details}
\label{supp:baseline_details}
For \textit{novel view video synthesis} (Table~\ref{tab:nvs-consistent4d},~\ref{tab:nvs-objaverse}, and~\ref{tab:nvs_anchor}), we compare SV4D with SV3D~\citep{voleti2024sv3d}, Diffusion$^2$~\citep{yang2024diffusion}, STAG4D~\citep{yang2024diffusion},
4Diffusion~\citep{zhang20244diffusion}. 
The results of SV3D, Diffusion$^2$, and 4Diffusion are generated with their official code.
STAG4D used Zero123++~\citep{shi2023zero123++} as the multi-view generator, which fixed the view angle, and hence the novel views generated from STAG4D cannot be changed to be consistent with the views evaluated.
We reproduced 
STAG4D with SV3D as the multi-view generator. SV3D has been shown to generate more consistent 3D results than Zero123++, so this serves as a stronger baseline.

For \textit{4D generation}, we compare SV4D with Consistent4D~\citep{jiang2023consistent4d}, STAG4D~\citep{zeng2024stag4d}, 
4Diffusion~\citep{zhang20244diffusion},
DreamGaussian4D (DG4D)~\citep{ren2023dreamgaussian4d}, GaussianFlow~\citep{gao2024gaussianflow}, 4DGen~\citep{yin20234dgen}, Efficient4D~\citep{pan2024fast}.
On the Consistent4D dataset (Table~\ref{tab:4d-consistent4d}), the results of Consistent4D, STAG4D, Efficient4D, and GaussianFlow are from the original paper.
4DGen results are from Efficient4D paper, and DG4D results are from GaussianFlow paper.
The results of 4Diffusion are generated with its official code.
On the ObjaverseDy dataset (Table~\ref{tab:4d-objaverse}), all results of our baselines are generated from their official code.

\section{Supplementary Video}
\label{supp:supp_video}
Beyond the paper, our supplementary materials offer a comprehensive video that provides an in-depth introduction to our task and method. Additionally, we add more visual comparisons which further showcase the effectiveness of our approach.

\end{document}